\documentclass[fleqn,10pt]{style/wlscirep}

\usepackage[utf8]{inputenc}
\usepackage[T1]{fontenc}
\usepackage{url}
\usepackage{amsmath,amsfonts}
\usepackage{pifont}
\usepackage{amssymb}
\usepackage{array}
\usepackage{multirow}
\usepackage{booktabs}
\usepackage{listings}
\usepackage{algorithm}
\usepackage{algorithmic}
\usepackage{graphicx}
\usepackage{stfloats}
\usepackage{color}
\usepackage{subcaption}
\usepackage{tabularx}
\usepackage{makecell}
\usepackage{verbatim}
\usepackage{textcomp}
\usepackage{lipsum}
\usepackage{ragged2e} % for \justifying
\newcommand{\ie}{i.e.,\ }
\newcommand{\eg}{e.g.,\ }
\newcommand{\aka}{a.k.a.\ }
\newcommand{\diff}{\mathrm{d}}
\newcolumntype{C}{>{\centering\arraybackslash}X}
\newcommand{\ExtraPercent}[1]{${\scriptstyle (+ #1 \%)}$}
\definecolor{CustomGreen}{RGB}{34,139,34}
\newcommand{\MyComment}[1]{\textcolor{CustomGreen}{\textit{// #1}}}

\DeclareMathAlphabet{\mathcal}{OMS}{cmsy}{m}{n} % use the decent mathcal font!!!

\title{
    Scalable Dendritic Modeling Advances Expressive and Robust Deep Spiking Neural Networks
}

\author[1,2]{Yifan Huang}
\author[3]{Wei Fang}
\author[2]{Zhengyu Ma}
\author[4,*]{Guoqi Li}
\author[1,2,3,*]{Yonghong Tian}

\affil[1]{School of Computer Science, Peking University, Beijing 100871, China}
\affil[2]{Peng Cheng Laboratory, Shenzhen 518000, China}
\affil[3]{School of Electronic and Computer Engineering, Shenzhen Graduate School, Peking University, Shenzhen 518055, China}
\affil[4]{Institute of Automation, Chinese Academy of Sciences, Beijing 100190, China}

\affil[*]{guoqi.li@ia.ac.cn, yhtian@pku.edu.cn}

\begin{abstract}
Dendritic computation endows biological neurons with rich nonlinear integration and high representational capacity, yet it is largely missing in existing deep spiking neural networks (SNNs). Although detailed multi-compartment models can capture dendritic computations, their high computational cost and limited flexibility make them impractical for deep learning. To combine the advantages of dendritic computation and deep network architectures for a powerful, flexible and efficient computational model, we propose the dendritic spiking neuron (DendSN). DendSN explicitly models dendritic morphology and nonlinear integration in a streamlined design, leading to substantially higher expressivity than point neurons and wide compatibility with modern deep SNN architectures. Leveraging the efficient formulation and high-performance Triton kernels, dendritic SNNs (DendSNNs) can be efficiently trained and easily scaled to deeper networks. Experiments show that DendSNNs consistently outperform conventional SNNs on classification tasks. Furthermore, inspired by dendritic modulation and synaptic clustering, we introduce the dendritic branch gating (DBG) algorithm for task-incremental learning, which effectively reduces inter-task interference. Additional evaluations show that DendSNNs exhibit superior robustness to noise and adversarial attacks, along with improved generalization in few-shot learning scenarios. Our work firstly demonstrates the possibility of training deep SNNs with multiple nonlinear dendritic branches, and comprehensively analyzes the impact of dendrite computation on representation learning across various machine learning settings, thereby offering a fresh perspective on advancing SNN design.
\end{abstract}

\begin{document}

% \linenumbers %! line numbering

\flushbottom
\maketitle
\thispagestyle{empty}

\section*{Introduction}

\begin{figure}[t!]
\centering
\includegraphics[width=\linewidth]{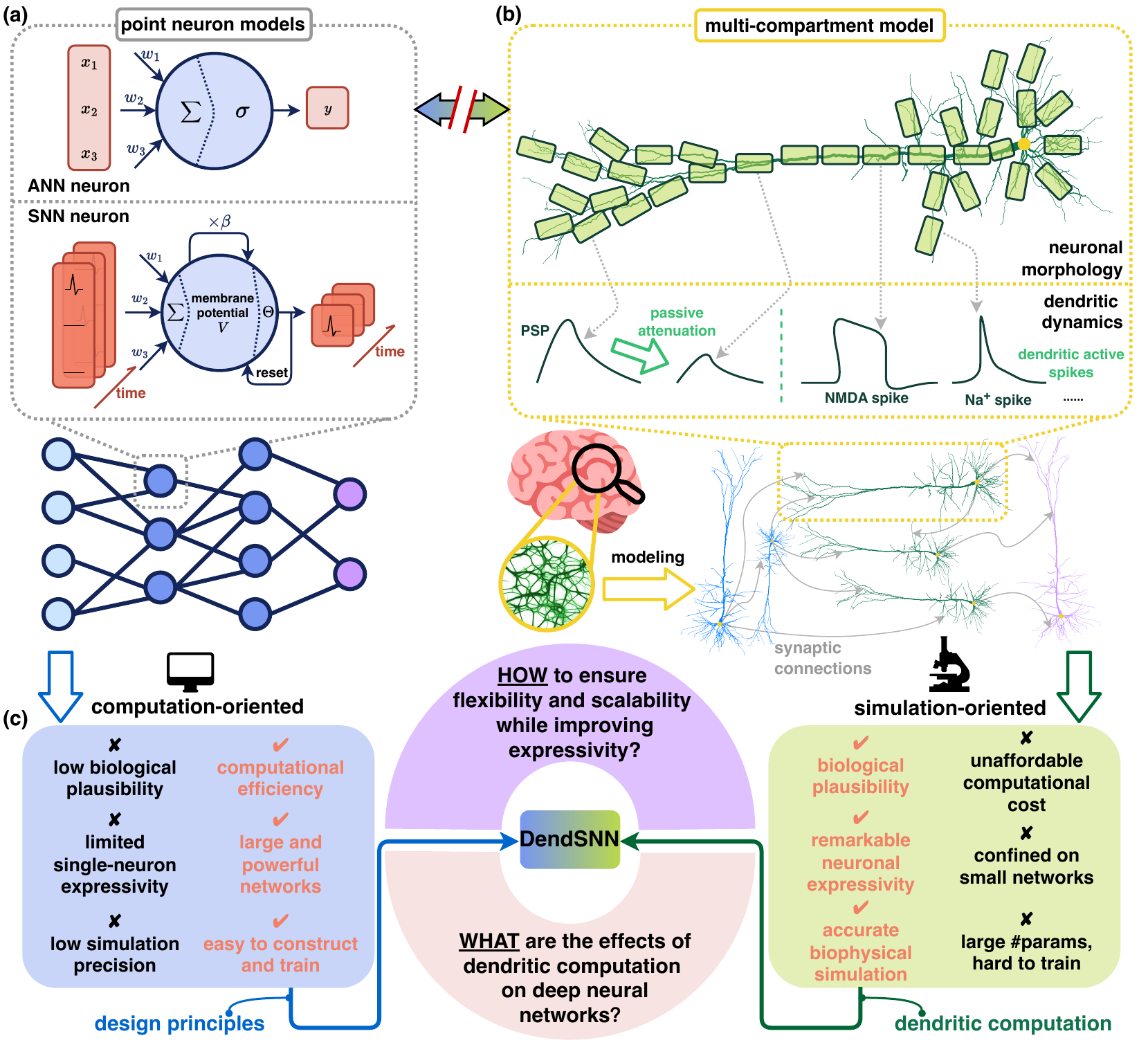}
\caption{
    \justifying
    \textbf{A comparison between deep neural networks with point neurons and biophysical neural networks with multi-compartment neurons.}
    \textbf{(a)} Deep neural networks typically adopt point neuron models. Both the perceptron in deep ANNs and the LIF model in deep SNNs belong to this category.
    \textbf{(b)} Multi-compartment models in neuroscience capture the detailed morphologies of biological neurons and have complex dendritic dynamics. By connecting these neurons, biophysical network models are established to simulate small-scale neural circuits in the brain.
    \textbf{(c)} Deep neural networks based on point neurons are computationally efficient and easy to train, but fall short on bio-plausibility and single-neuron expressivity (blue box). Biophysical networks with multi-compartment neurons have remarkable neuron-level expressivity, but cannot be easily scaled up to large networks (green box). There is a gap between these two types of models. In this work, we propose DendSN and DendSNN, combining dendritic computation with the design principles of deep learning to power up deep SNNs.
}
\label{fig:introduction}
\end{figure}

The past decade has witnessed the success of deep learning across diverse domains, including computer vision \cite{krizhevsky2017alexnet,he2016resnet,dosovitskiy2021vit}, natural language processing \cite{vaswani2017transformer,devlin2019bert,achiam2023gpt4,guo2025deepseek}, and autonomous driving \cite{lechner2020ncp,vorbach2021causal}. Fueled by the rapid progress of parallel computing devices like GPUs, increasingly deeper artificial neural networks (ANNs) can be efficiently trained and deployed for real-world applications \cite{coates2013hpc,chetlur2014cudnn}. In addition, inspired by the information processing mechanisms of biological neural circuits, spiking neural networks (SNNs) have emerged as a potentially more bio-plausible and energy-efficient alternative to ANNs \cite{hassabis2017neuroscience,roy2019towards}. Recent advances in neuromorphic hardware \cite{davies2018loihi,debole2019truenorth,pei2019tianjic,yao2024speck} and SNN programming frameworks \cite{hazan2018bindsnet,eshraghian2023snntorch,fang2023spikingjelly} have further accelerated the development of deep SNNs, positioning them as promising models for the next generation of neural networks \cite{maass1997networks}.

The design of deep SNNs typically follows two complementary directions: network architectures and neuron models. On the architectural side, recent breakthroughs have adapted components from ANNs, such as residual connections \cite{he2016resnet} and self-attention \cite{vaswani2017transformer}, to their spiking counterparts, leading to notable performance improvements. On the neuronal side, inspirations can arise from mathematical abstraction \cite{fang2020plif,yin2023fptt,huang2024clif,stanojevic20240.3spikes} and engineering perspectives \cite{fang2023psn,li2024psu}. What's more, neuroscience studies have revealed that the computational capacity of a biological neuron rivals that of a multi-layer neural network consisting of thousands of artificial neurons \cite{poirazi2003pyramidal,jadi2014twolayer,tzilivaki2018dogma,beniaguev2021cortical}. Also, replicating the dynamics of a single bio-plausible Hodgkin-Huxley (HH) model needs four simple LIF neurons \cite{he2024hh}. These findings highlight that biological neurons possess considerably higher internal complexity than their artificial counterparts used in deep networks. Incorporating bio-inspired mechanisms into neuron modeling is therefore a promising approach to enhance the expressivity of deep SNNs \cite{hammouamri2022threshold,hammouamri2024delay,he2024hh}. Yet, what mechanisms can be abstracted and how they can be effectively utilized for improved performance in practical tasks remain open questions.

The dynamics of biological neurons originate mainly from two sources: dendrites and soma. Most existing deep SNNs, however, are built upon oversimplified neurons such as the leaky integrate-and-fire (LIF) model \cite{lapicque1907lif}, which treats the entire neuron as a single spatial point and depicts only somatic dynamics. Models like the parametric LIF (PLIF) \cite{fang2020plif} and few-spikes neuron (FS-neuron) \cite{stockl2021optimized} enhance neuronal dynamics by introducing learnable parameters, but still ignore dendritic computation. These spiking models, as well as their non-spiking counterparts in ANNs \cite{sanger1958perceptron}, are termed \textbf{point neurons} due to their reduced morphology and dynamics (Figure~\ref{fig:introduction}(a)). Despite their efficiency, point neurons sacrifice essential computational properties of dendrites, leading to limited single-neuron expressivity. This oversimplification poses a potential bottleneck for deep SNNs' computational capacity (Figure~\ref{fig:introduction}(c), blue box).

In contrast, neuroscientists employ interconnected \textbf{multi-compartment models} to simulate the biophysical activities of small-scale neural circuits \cite{hines2001NEURON,stimberg2019Brian2}. These models describe detailed neuronal morphology and use coupled differential equations to capture both passive and active dendritic dynamics \cite{poirazi2003arithmetic,schutter1994purkinje,hay2011pyramidal} (Figure~\ref{fig:introduction}(b)). They can reproduce critical dendritic functions \cite{london2005dendritic,payeur2019dendriticinfo,acharya2022dendritic} including passive signal attenuation \cite{mengual2020dendsoma}, local dendritic spikes \cite{major2013active}, and input selection or multiplexing \cite{payeur2019dendriticinfo}. As a result, multi-compartment models exhibit greater neuron-level capacity (Figure~\ref{fig:introduction}(c), green box). Extending deep SNNs with structural and dynamical dendritic mechanisms is thus an intuitive strategy for improving network-level performance. This direction, however, has not been fully explored in deep SNN research.

Incorporating dendritic computation into deep SNNs presents critical challenges in terms of scalability and flexibility. The success of deep SNNs depends on their large scales and intricate architectures, yet simulating large populations of multi-compartment neurons with complex dynamics remains computationally prohibitive \cite{guerguiev2017towards,sacramento2018microcircuits,payeur2021burst}. Even with simplifications, such neurons are often not trivially compatible with modern network components such as convolution \cite{zheng2024temporal,plagge2024expressive,wu2018improved}. To ensure scalability and applicability of dendritic deep SNNs (Figure~\ref{fig:introduction}(c), upper part of the ring), efficient and flexible neuron population modeling as well as a comprehensive exploitation of GPU acceleration are required. Meanwhile, although dendritic computation has been extensively investigated in neuroscience \cite{london2005dendritic,hay2011pyramidal,payeur2019dendriticinfo,li2019dendritic,bicknell2021synaptic}, its role in deep SNNs remains limited to specific scenarios such as sequence modeling \cite{zheng2024temporal,chen2024pmsn,zhang2024tclif,wang2025mmdend}. A systematic understanding of how dendritic computation affects representation learning and task performance across multiple scenarios is still lacking (Figure~\ref{fig:introduction}(c), lower part of the ring).

To address these challenges, we propose the dendritic spiking neuron (DendSN) that explicitly models dendritic morphology and nonlinear dynamics in a lightweight manner. We demonstrate that DendSN has significantly higher expressivity than point spiking neurons due to its nonlinear dendritic aggregation mechanism. At the network level, DendSNs can be seamlessly integrated into various deep SNN architectures, showcasing their flexibility. We further develop efficient Triton kernels \cite{tillet2019triton} to fully leverage GPU parallelism, enabling deep dendritic spiking networks (DendSNNs) to scale to depths comparable to traditional SNNs while keeping computational costs affordable. Experiments show that DendSNNs consistently outperform SNNs based on point neurons across static and event-based tasks with negligible increases in parameter counts. At the application level, we comprehensively evaluate DendSNNs' performances across various machine learning settings to show their broader benefits. Inspired by dendritic modulation \cite{wybo2023NMDA} and synaptic clustering \cite{cichon2015branch,limbacher2020cluster} in biological neural circuits, we propose dendritic branch gating (DBG) algorithm for task-incremental learning, which effectively mitigates interference between different tasks in DendSNNs. Additional results show that DendSN can enhance SNNs' robustness to noise and adversarial attacks, and improve few-shot learning performance. To the best of our knowledge, this is the first work to construct and train deep SNNs with multiple nonlinear dendritic branches and to systematically analyze their advantages in diverse machine learning scenarios.

\section*{Results}
\label{sec:results}

\subsection*{Balancing expressivity and computational cost of dendritic spiking neuron model}
\label{subsec:results-dendsn}

\begin{figure}[t!]
\centering
\includegraphics[width=\linewidth]{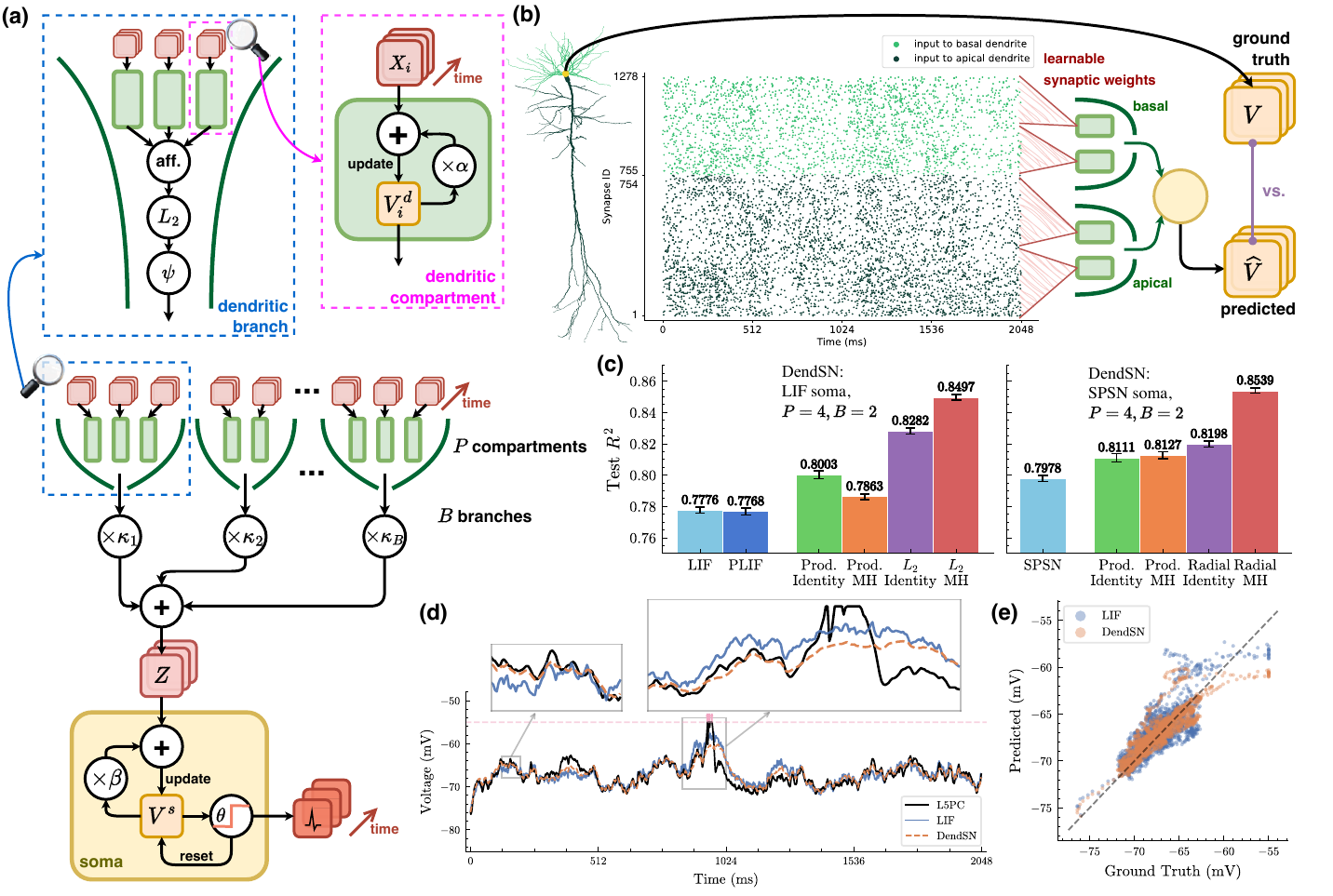}
\caption{
    \justifying
    \textbf{DendSN and L5PC fitting task.}
    \textbf{(a)} The proposed DendSN model with stateful dendrite and LIF soma.
    \textbf{(b)} The setting of the L5PC somatic potential prediction task.
    \textbf{(c)} The coefficient of determination ($R^2$) of different neuron models on L5PC somatic potential prediction. Higher is better.
    \textbf{(d),(e)} The somatic membrane potential of a detailed multi-compartment L5PC model (black, ground truth), a LIF model (blue), and a DendSN (orange and dashed, $P=4,B=2$). The trial is selected from the test set.
}
\label{fig:dendsn}
\end{figure}

Multi-compartment models in computational neuroscience depict neuronal morphology at a micrometer level and portray dynamics using an extensive set of variables  \cite{schutter1994purkinje,poirazi2003arithmetic,hay2011pyramidal}, hence being able to capture complex activity patterns of biological neurons (Figure~\ref{fig:introduction}(b)). Nevertheless, their high computational cost impedes their application in deep neural networks and complex tasks, even with optimal acceleration methods in place \cite{zhang2023gpubased}. A key challenge is therefore to simplify the neuron model while preserving essential structural and computational properties.

We propose the dendritic spiking neuron (DendSN) model illustrated in Figure~\ref{fig:dendsn}(a) as a possible solution to the problem. The model runs on $T$ discrete time steps, following the standard practice of SNNs \cite{fang2023spikingjelly}. Structurally, a DenSN consists of $P$ dendritic compartments distributed over $B$ segregated branches and a soma. The dendritic morphology is simplified by considering only which branch each compartment belongs to, while neglecting the fine-grained connectivity among compartments within individual branches. In terms of dynamics, each dendritic compartment $i$ acts as a leaky integrator with a state $V^d_i$ and a shared decay factor $\alpha$ (Figure~\ref{fig:dendsn}(a), pink box). At the branch level, the instantaneous states of all compartments on branch $b$ are first affinely transformed and then aggregated through $L_2$ norm, followed by a dendritic activation $\psi$ (Figure~\ref{fig:dendsn}(a), blue box). The somatic input is the weighted sum of the outputs from all the branches. The soma itself can be instantiated as any point spiking neuron model. Typical choices are the integrate-and-fire (LIF) neuron \cite{lapicque1907lif,gerstner2014neuronal}, the parametric LIF (PLIF) neuron \cite{fang2020plif}, and the sliding parallel spiking neuron (SPSN) \cite{fang2023psn}. See Methods for a formal definition of DendSN.

Compared to point neurons, DendSN's enhanced expressivity arises from two key mechanisms. First, each dendritic compartment implements a leaky integrator, so the local states $\{V^d_i[t]\}_{i=1}^P$ act as temporally filtered features that accumulate and retain past synaptic input. This intrinsic temporal filtering increases the neuron's memory capacity \cite{zheng2024temporal} and allows it to capture long-term temporal dependencies. Second, the instantaneous mapping from compartment states to the somatic input $Z[t]$ can be formulated as
\begin{equation}
\label{eq:results-dendrite}
Z[t] = \sum_{b=1}^B \kappa_b \psi\left\{ \left[ \sum_{i \in C_b} \left(\frac{V_i^d[t] - \xi_i}{\zeta_b}\right)^2\right]^{\frac{1}{2}} \right\},
\end{equation}
where $\kappa_b$ is the weighting factor for branch $b$, $\xi_i$ and $\zeta_b$ are the translation and dilation factors for the affine transformation, and $C_b$ denotes the set of compartments on branch $b$ (see Methods). When the dendritic activation function $\psi$ is chosen as a standard wavelet (\eg the Mexican hat wavelet), the mapping in Equation~\eqref{eq:results-dendrite} is mathematically analogous to a wavelet neural network (WNN) \cite{zhang1995wnn,alexandridis2013wnnguide,liu2021wcnn} with the following form:
\begin{equation}
\label{eq:results-wnn}
y = \sum_{b=1}^B \kappa_b \psi\left\{ \left[ \sum_{i=1}^{P} \left(\frac{x_i - \xi_i}{\zeta_b}\right)^2\right]^{\frac{1}{2}} \right\}.
\end{equation}
In this sense, the dendrite can be viewed as a WNN with $P$ inputs and $B$ hidden units, whose first layer is block-sparse. Together, the temporal filtering and the WNN-like nonlinear projection enable DendSN to extract richer spatiotemporal features than point neurons, while retaining an architecture amenable to deep learning implementations.

To validate DendSN's expressivity, we fit it to activity data collected from a multi-compartment model of a layer-5 pyramidal cell (L5PC), as shown in Figure~\ref{fig:dendsn}(b). For details of the experiment, refer to Methods. By leveraging gradient descent to learn synaptic weights $\{W_j\}_{j=1}^{N}$ ($N=1278$ is the number of synaptic channels, and $W_j\in \mathbb{R}$) as well as neuronal parameters $\alpha$,$\{\xi_i\}_{i=1}^P$,$\{\zeta_b\}_{b=1}^B$ and $\{\kappa_b\}_{b=1}^{B}$, a standard DendSN with $P=4$ compartments, $B=2$ branches, $L_2$ branch aggregation, Mexican hat dendritic activation and LIF soma can model the mapping from presynaptic spikes to somatic membrane potential of the detailed biophysical model with high fidelity. A coefficient of determination ($R^2$, higher is better) of $0.8497$ is achieved on the test set (Figure~\ref{fig:dendsn}(c), left, red), which is significantly higher than that of LIF ($R^2=0.7776$) and PLIF ($R^2=0.7768$). The DendSN with the same dendrite configuration and a SPSN soma yields an even higher $R^2$ of $0.8539$, mainly thanks to the stronger sequence modeling ability of SPSN compared to LIF and PLIF. That also outperforms a bare SPSN ($R^2=0.7978$). As shown in Figure~\ref{fig:dendsn}(d) and Figure~\ref{fig:dendsn}(e), although all reduced models fail to predict the potential accurately near spike onset, the somatic potential curve given by DendSN (orange, dashed) better tracks the ground truth (black) within the subthreshold regime compared to that given by LIF (blue). Moreover, the prediction given by DendSN is smoother than that produced by LIF, reflecting the low-pass filtering effect of the dendrite (see Supplementary Materials~\ref{supsec:low-pass-filter} for details). Importantly, the additional number of parameters from the dendrites is negligible compared to that from synaptic weights (9 dendritic parameters vs. 1278 synaptic parameters), indicating that the performance gains are due to architectural expressivity rather than parameter count.

We further investigate the effect of DendSN's WNN-like dendritic mapping on its expressivity. Replacing the Mexican hat activation with an identity mapping significantly decreases performance (Figure~\ref{fig:dendsn}(c), purple), highlighting the importance of dendritic nonlinearity. Altering branch aggregation rule from $L_2$ norm to product, as done in some WNN implementations \cite{alexandridis2013wnnguide}, also reduces performance (Figure~\ref{fig:dendsn}(c), orange). This suggests that product-based aggregation may exacerbate gradient vanishing and impede learning. Detailed definitions of these DendSN variants can be found in Methods. Overall, with proper design choices, DendSN turns out to possess solid biological plausibility and high expressivity.

\subsection*{Constructing deep dendritic spiking neural networks}
\label{subsec:results-dendsnn}

\begin{figure}[t!]
\centering
\includegraphics[width=\linewidth]{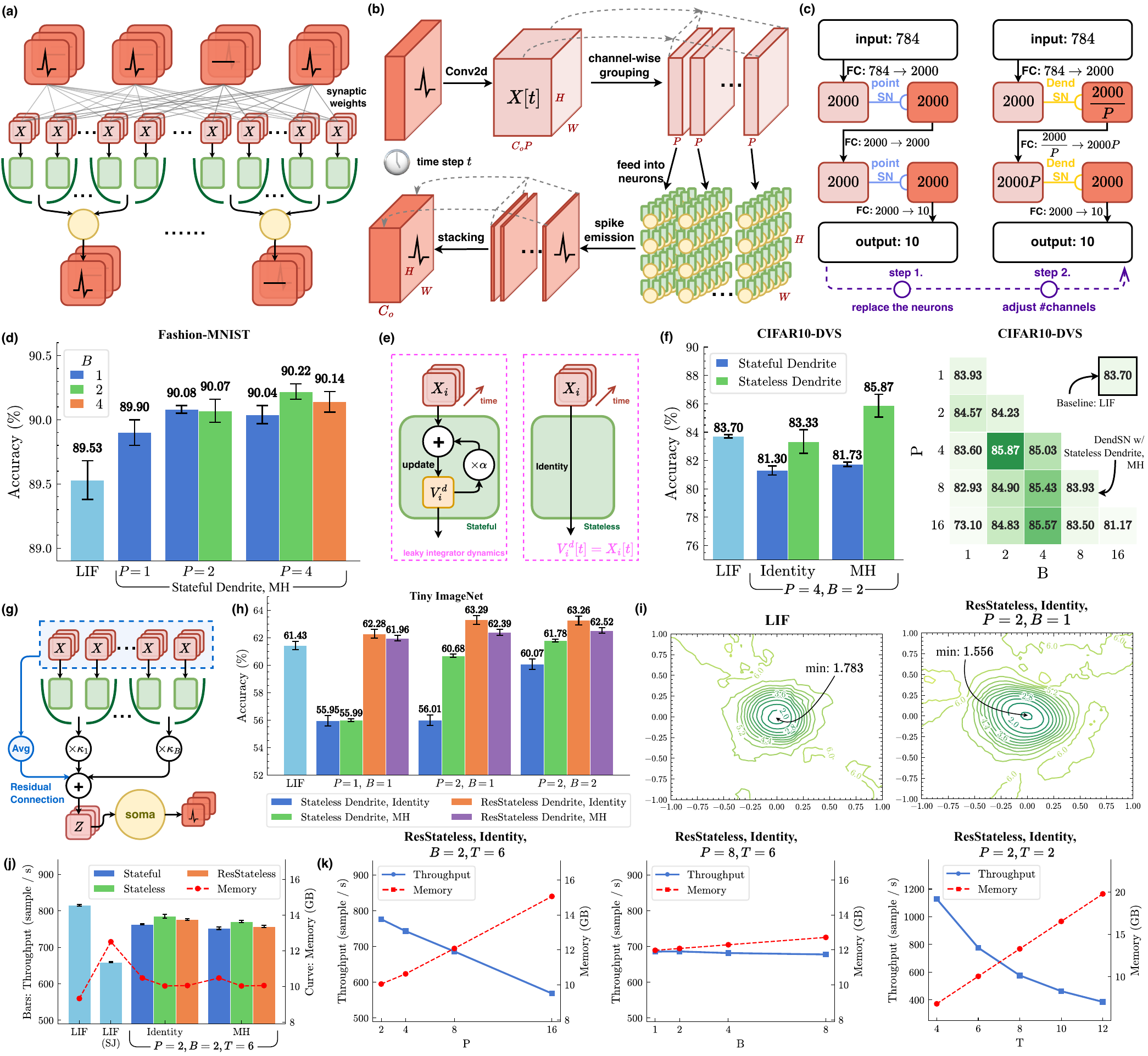}
\caption{
    \justifying
    \textbf{Construction of DendSNNs and their evaluation on classification tasks.}
    \textbf{(a)} A DendSN layer placed after a fully connected layer.
    \textbf{(b)} A DendSN layer placed after a 2D convolutional layer.
    \textbf{(c)} Deriving a DendSNN architecture from a PointSNN by replacing the neurons and adjusting the number of channels.
    \textbf{(d)} Accuracy comparison on Fashion-MNIST.
    \textbf{(e)} Stateful and stateless dendritic compartments.
    \textbf{(f)} Left: accuracy comparison between Stateful and Stateless Dendrites on CIFAR10-DVS. Right: CIFAR10-DVS accuracies under different $P$ and $B$ settings.
    \textbf{(g)} DendSN with dendritic residual connection.
    \textbf{(h)} Accuracy comparison on Tiny ImageNet.
    \textbf{(i)} Loss landscape visualizations of the models on Tiny ImageNet.
    \textbf{(j)} Comparison of different models' throughput and peak allocated memory when trained on Tiny ImageNet.
    \textbf{(k)} Evolution of training throughput and memory cost when $P$ (left), $B$ (mid) or $T$ (right) increases.
}
\label{fig:dendsnn}
\end{figure}

The success of modern deep learning largely stems from well-designed architectures and large network scales. Deep networks like ResNet \cite{he2016resnet} and Transformer \cite{vaswani2017transformer} have been widely adopted for tackling machine learning problems across diverse domains. Recent advances in SNN have built up spiking counterparts of these models, outperforming those shallow SNNs \cite{zheng2020going,fang2021sew,hu2024msresnet,zhou2023spikformer,yao2023spikedriventransformer,yao2024meta,zhou2024qkformer}. Here we incorporate DendSNs into various deep SNN architectures to jointly exploit the expressive dendritic computation and the representational capacity of deep networks.

To embed DendSNs into deep SNNs, neurons are organized as layers to facilitate tensor-based formulation. For simplicity, all DendSNs within a layer share the same structural configuration ($P$ and $B$), and each dendritic branch contains an equal number of compartments ($P/B$). We also assume that neurons on different channels have independent learnable parameters, while those at different spatial positions within the same channel share parameters; an exception is the compartmental decay factor $\alpha$, which is always shared by all neurons in the layer, following the practice of PLIF \cite{fang2020plif}. Each DendSN layer is placed after a weight layer. The channel dimension of the weight layer's output is factorized as $C = C_0 \times P$, where $C_0$ corresponds to DendSN layer's output channels and $P$ indexes dendritic compartments. Each group of $P$ consecutive feature maps is assigned to a neuron channel consisting of DendSNs with $P$ compartments (Figure~\ref{fig:dendsnn}(b)). As a result, a DendSN layer reduces the channel dimension by a factor of $P$ without altering spatial resolution. Figure~\ref{fig:dendsnn}(a) and Figure~\ref{fig:dendsnn}(b) illustrate the cases for fully connected and 2D convolutional layers, respectively. See Methods for more details.

By stacking multiple weight–DendSN blocks, deep dendritic spiking neural networks (DendSNNs) can be constructed. A DendSNN architecture can be intuitively derived from conventional point-neuron-based SNNs (PointSNN) by first replacing the point neuron layers with DendSN layers, and then adjusting the number of channels in the weight layers (Figure~\ref{fig:dendsnn}(c)). Notice that this conversion does not significantly increase the parameter count (see Supplementary Materials~\ref{supsec:parameter-counts}), making it reasonable to directly compare DendSNNs with their PointSNN counterparts. Backpropagation through time (BPTT) with surrogate gradient \cite{wu2018stbp,zenke2018superspike,shrestha2018slayer} is used to train DendSNNs end-to-end (detailed in Methods).

The formulation of DendSNN provides flexibility to balance three key aspects of deep SNNs: temporal memory capacity, computational efficiency, and ease of training. Correspondingly, we present three dendrite model variants to enable controlled tradeoffs. \textbf{Stateful Dendrite}, the original model defined in Equation~\eqref{eq:methods-compartment} and Equation~\eqref{eq:methods-branch-aggregation}, retains leaky-integrator–based compartmental dynamics to enhance temporal memory, making it well suited for small-scale tasks or tasks with rich temporal structure. \textbf{Stateless Dendrite} replaces compartmental dynamics with a simple identity mapping $V^d_i[t] = X_i[t]$ (Figure~\ref{fig:dendsnn}(e)), improving computational efficiency and making it suitable for tasks with weak temporal dependencies. \textbf{ResStateless Dendrite} further adds a residual connection from synaptic inputs $\{X_i[t]\}_{i=1}^P$ to the somatic afferent signal $Z[t]$ (Figure~\ref{fig:dendsnn}(g) and Equation~\eqref{eq:methods-residual}) to mitigate gradient vanishing and enable training of very deep DendSNNs. By selecting the dendrite variant that matches the task demand, DendSNNs can achieve superior performance while maintaining efficient training.

Classification experiments are conducted on static image and neuromorphic vision datasets to validate the effectiveness of DendSNNs, with detailed settings provided in Methods. On Fashion-MNIST \cite{xiao2017fashionmnist}, we use Stateful Dendrite, as the dataset is small and computational efficiency is not a primary concern. As Figure~\ref{fig:dendsnn}(d) shows, fully connected DendSNNs consistently outperform their PointSNN counterparts. Increasing the number of compartments $P$ generally improves classification accuracy, while moderate branch numbers $B$ yield better performance. 
For CIFAR10-DVS \cite{li2017cifar10dvs}, an event-based classification benchmark with weak temporal dependencies, we prefer Stateless Dendrite to reduce computational overhead. As the bar plot in Figure~\ref{fig:dendsnn}(f) shows, VGG-based DendSNNs with Stateless Dendrite achieve higher accuracies than those using Stateful Dendrites. A plausible explanation is that the removal of compartmental dynamics reduces the excessive temporal smoothing and thus leads to more stable optimization behavior on this dataset. When adopting the Mexican hat dendritic activation, DendSNNs with Stateless Dendrites ($P=4, B=2$) significantly outperform LIF-based SNNs. The heat map in Figure~\ref{fig:dendsnn}(f) further confirms that moderate $P$ and $B$ values yield optimal performance.
On Tiny ImageNet, training stability becomes critical. We therefore employ ResStateless Dendrite to facilitate effective optimization. As Figure~\ref{fig:dendsnn}(h) shows, DendSNNs with Stateless Dendrites fail to match the performance of LIF-based SNNs. In contrast, DendSNNs with ResStateless Dendrites surpass LIF-based SNNs when using identity branch activation, indicating that the residual pathway effectively alleviates training difficulties in deep DendSNNs. Visualization of the loss landscape in Figure~\ref{fig:dendsnn}(i) reveals that ResStateless Dendrite enables convergence to lower and flatter minima, supporting better generalization. Collectively, these results demonstrate that DendSNNs can effectively exploit dendritic computation to enhance classification performance.

While DendSNNs incorporate more sophisticated neuron models, the additional computational cost is marginal. The dendrite module is inherently parallelizable across compartments and branches, and its stateless variants further enable temporal parallelism. To convert this algorithmic parallelism into practical speedups, we develop Triton kernels for the forward and backward pass of dendritic integration (Equation~\eqref{eq:results-dendrite}), achieving high GPU utilization. As shown in Figure~\ref{fig:dendsnn}(j), on the network for Tiny ImageNet, DendSNNs maintain approximately $0.92 \times$ the throughput of LIF-based SNNs implemented with Triton, while greatly surpassing the CuPy-based LIF implementations provided by SpikingJelly \cite{fang2023spikingjelly}. In terms of memory efficiency, we apply gradient checkpointing \cite{chen2016checkpointing} to the dendritic integration process, which recomputes intermediate results during backward pass instead of saving them during forward pass to reduce memory cost. This strategy leads to only about $12\%$ additional memory cost compared to LIF-based SNNs implemented with Triton, while still remaining more memory efficient than SpikingJelly's LIF implementation (Figure~\ref{fig:dendsnn}(f), red curve). Among the three dendrite variants, Stateless Dendrite exhibits the highest efficiency, followed by ResStateless Dendrite, with Stateful Dendrite being the least efficient. The choice of branch activation $\psi$, however, has negligible influence on throughput or memory. Also, as shown in Figure~\ref{fig:dendsnn}(k), the number of compartments $P$ and time steps $T$ significantly affect speed and memory cost, while the number of branches $B$ has a relatively limited effect. Overall, these results highlight that DendSNNs preserve high efficiency and scalability, making large-scale dendritic SNNs practically feasible for deep learning applications.

\subsection*{Task Incremental Learning via Dendritic Modulation}
\label{subsec:results-dbg}

\begin{figure}[t!]
\centering
\includegraphics[width=\linewidth]{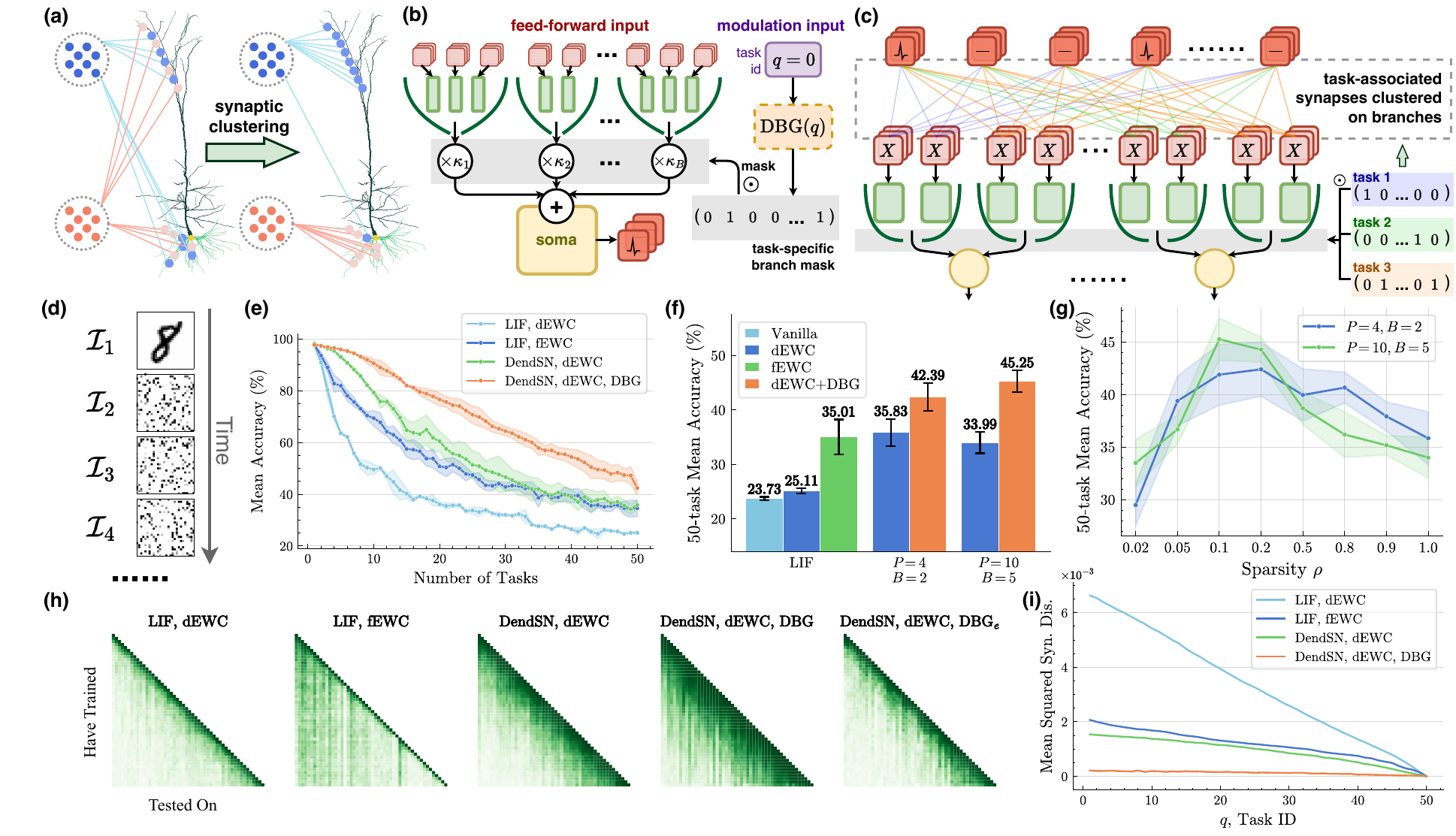}
\caption{
    \justifying
    \textbf{Dendritic branch gating (DBG) for task-incremental learning (TIL).}
    \textbf{(a)} Synaptic clustering on biological dendrites.
    \textbf{(b)} An illustration of DBG on a single DendSN.
    \textbf{(c)} DBG induces task-specific synaptic clusters in DendSNNs.
    \textbf{(d)} An illustration of the Permuted MNIST TIL benchmark.
    \textbf{(e)} Evolution of mean accuracy for different models on Permuted MNIST.
    \textbf{(f)} 50-task mean accuracies of different models on Permuted MNIST.
    \textbf{(g)} The effect of sparsity factor $\rho$ on TIL performance.
    \textbf{(h)} Accuracy heatmaps. The pixel on row $i$ and column $j$ represents the test accuracy of task $i$ after training on task $1$ to task $j$. Darker colors indicate higher accuracies. The wider the dark region, the less the model forgets across tasks.
    \textbf{(i)} Mean squared synaptic distance between the first-layer weights after training on task $q$ and those of the final network.
    dEWC and fEWC denote EWC applied to the decoder and full network, respectively.
    Unless otherwise stated, DendSNs use $P=4$, $B=2$, Stateful Dendrite, and Identity branch activation.
}
\label{fig:dbg}
\end{figure}

Having introduced DendSNN's formulation and its remarkable expressivity, we now examine the impact of dendritic computation in more challenging machine learning scenarios. We first consider task incremental learning (TIL) \cite{wang2023continual}, where a model must adapt to new tasks without forgetting previously learned knowledge (see Methods for a formal definition). This ability is inherent in biological intelligence, but poses a major challenge for neural networks \cite{goodfellow2013empirical}. We hypothesize that incorporating dendritic computation can help preserve previously acquired knowledge.

In the brain, sensory neurons receive not only bottom-up feedforward inputs but also top-down modulatory signals that adjust neuronal responses \cite{roth2015thalamic}. These modulation inputs typically originate from motor and prefrontal areas and convey higher-level information such as task context \cite{atiani2009task,popovkina2022task}. Wybo et al. \cite{wybo2023NMDA} revealed that NMDA-driven dendritic spikes may underlie contextual modulation of hierarchical sensory pathways, highlighting a potential mechanism for TIL.

In addition, the spatial organization of synaptic sites on dendrites strongly influences neuronal responses \cite{mel1994information,bono2017modelling}. A pertinent hypothesis is synaptic clustering, which posits that functionally related synapses tend to cluster on dendritic branches as a result of structural plasticity (Figure~\ref{fig:dbg}(a)). This mechanism enhances the brain's memory capacity by enabling relevant features to be preprocessed locally on dendritic tree before somatic integration, thus reducing interference from irrelevant signals \cite{kastellakis2015synaptic}. Evidence from both anatomy and computational modeling supports this view \cite{takahashi2012locally,cichon2015branch,limbacher2020cluster}.

Inspired by dendritic modulation and synaptic clustering, we propose a novel algorithm named dendritic branch gating (DBG) to mitigate catastrophic forgetting of DendSNNs in TIL scenarios (Figure \ref{fig:dbg}(b)). The index of the current task, denoted as $q$, serves as a top-down modulation signal and is fed to the network alongside the feedforward input during both training and inference. Note that the use of an extra contextual signal is a common practice in previously proposed TIL algorithms \cite{masse2018xdg}. For each DendSN layer in the network, DBG generates a sparse binary dendritic branch mask based on $q$ and applies it by element-wise multiplication during both training and inference (Equation~\eqref{eq:methods-dbg-mask}). Each element of the mask is independently sampled from a Bernoulli distribution with parameter $\rho$ (Equation~\eqref{eq:methods-dbg-sampling}), following the practice of context-dependent gating (XdG) \cite{masse2018xdg}. Notice that the unmasked branch weights are learnable, while the mask is always fixed. This intuitive approach has distinct functional effects. Training-time masking promotes the formation of task-specific synaptic clusters by restricting each task's updates to a small subset of dendritic branches (Figure~\ref{fig:dbg}(c)). Inference-time masking, on the other hand, ensures the activation of the correct subnetwork for a given task. Refer to Methods for more details about DBG. 

In contrast to XdG \cite{masse2018xdg}, which applies task-specific masks to neuron outputs, DBG performs gating at the dendritic branch. This finer granularity allows a neuron to participate in multiple tasks through different combinations of its dendritic branches, providing a substantially larger combinatorial capacity than XdG and enabling more flexible task-specific subnetwork allocation.

To evaluate TIL performance, we train fully connected networks on the Permuted MNIST benchmark \cite{goodfellow2013empirical,masse2018xdg} consisting of $50$ tasks (Figure~\ref{fig:dbg}(d)). Experimental details are provided in Methods. As shown in Figure~\ref{fig:dbg}(f), the LIF-based SNN exhibits severe catastrophic forgetting, yielding a mean accuracy of only $23.73\%$ across 50 tasks. Applying elastic weight consolidation (EWC) \cite{kirkpatrick2017overcoming} to the decoder (dEWC) fails to significantly improve performance, while applying EWC to the full network (fEWC) increases 50-task mean accuracy to $35.01\%$. In comparison, DendSNNs (Stateful Dendrite, Identity) combined with dEWC achieve a comparable 50-task mean accuracy to LIF-based SNNs with fEWC. As Figure~\ref{fig:dbg}(e) demonstrates, DendSNNs with dEWC perform even better than LIF-based SNNs with fEWC in the early stage ($<30$ learned tasks). Incorporating DBG into DendSNNs with dEWC further raises the mean accuracy to above $42\%$, significantly outperforming the best LIF-based counterpart with fEWC (Figure~\ref{fig:dbg}(f)). The performance gain arises because DBG's sparse gating directs the parameter updates of different tasks to largely separate parameter subspaces, reducing interference more effectively than EWC's soft regularization-based constraint. A sweep of sparsity factor $\rho$ in Figure~\ref{fig:dbg}(g) reveals that a moderate $\rho$ value leads to the best performance ($\rho=0.2$ for $P=4, B=2$; $\rho=0.1$ for $P=10, B=5$). A lower $\rho$ value makes the parameter subspaces more disjoint, reducing inter-task interference. However, an extremely low $\rho$ value reduces subnetwork capacity and hinders the learning of new tasks. Thus, a moderate $\rho$ achieves the best balance between task separation and per-task representational capacity.

To gain deeper insights into how DendSNNs and DBG mitigate catastrophic forgetting, we visualize the accuracy matrices in Figure~\ref{fig:dbg}(h). The element at row $i$ and column $j$ represents the test accuracy of task $i$ after the first $j$ tasks are learned, with darker colors indicating higher accuracies. The gradual fading from the diagonal to the lower-left corner reflects the progressive forgetting of earlier tasks. With fEWC, LIF-based SNNs can partly preserve the performance of early tasks even after many subsequent tasks are learned. Replacing LIF with DendSN while applying EWC only to the decoder enhances the memorization of recently learned tasks, though performance of tasks in the distant past still declines. Incorporating DBG further extends the retention window, suggesting improved memory capacity that arises from synaptic clustering. We also introduce a dense variant of DBG called DBG-embedding ($\text{DBG}_e$), which retains dendritic modulation but removes sparse branch connectivity (see Methods). As shown in Figure~\ref{fig:dbg}(h), $\text{DBG}_e$ performs worse than DBG and even than DendSNNs without dendritic modulation, confirming that branch sparsity is crucial for synaptic clustering and continual learning. Figure~\ref{fig:dbg}(i) further demonstrates the mean squared synaptic distance between the first-layer weights after training on task $q$ and those after training on all $Q$ tasks (detailed in Methods). Both DendSN and DBG can effectively suppress synaptic drift when learning on subsequent tasks, which is the underlying reason for their improved performance. Together, these results demonstrate that DendSN and DBG jointly stabilize representations and substantially alleviate catastrophic forgetting in TIL.

\subsection*{Robustness against noise and adversarial attacks}
\label{subsec:results-robustness}

\begin{figure}[t!]
\centering
\includegraphics[width=\linewidth]{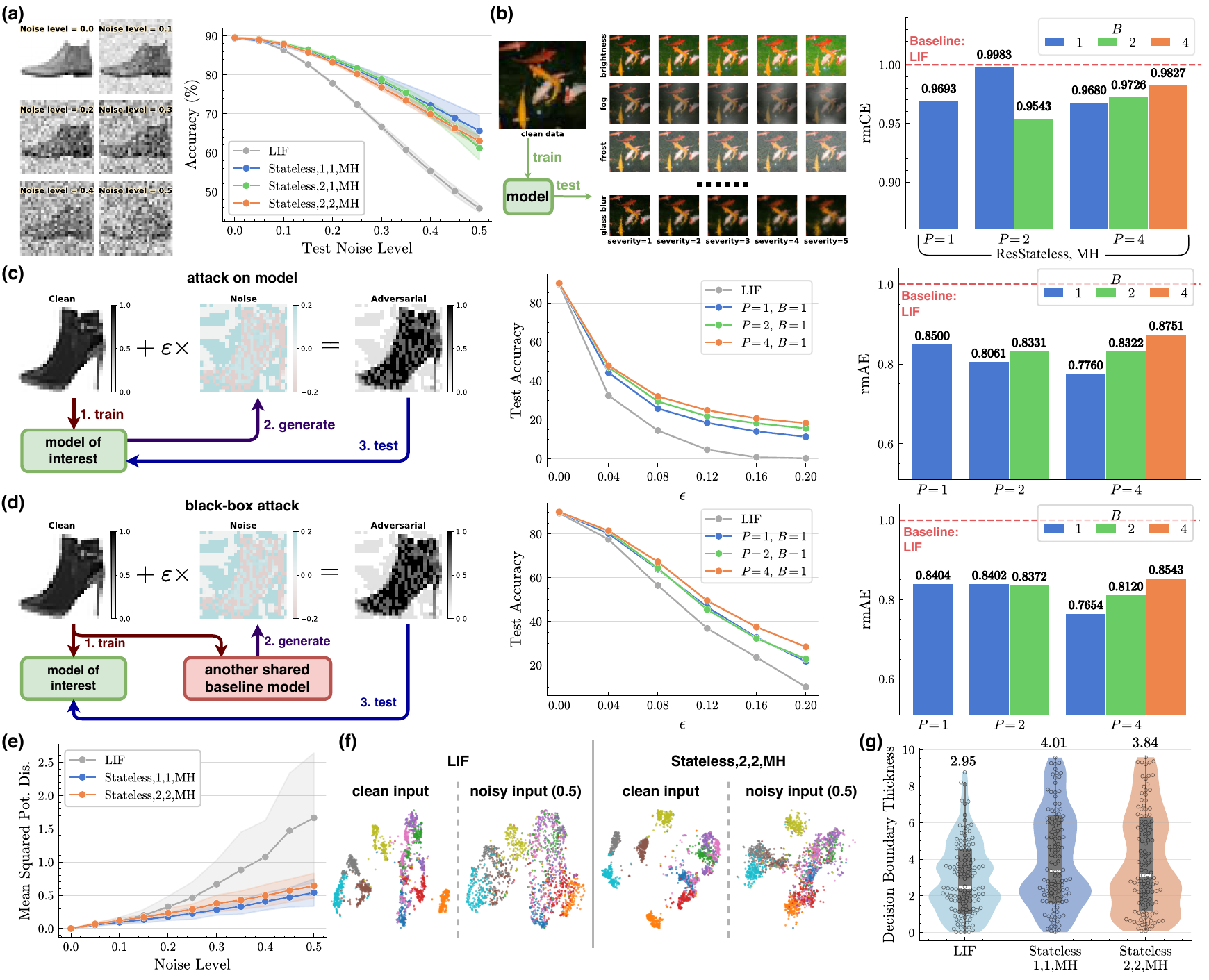}
\caption{
    \justifying
    \textbf{Robustness of DendSNNs against noise and adversarial attacks.}
    \textbf{(a)} Fashion-MNIST noise robustness experiment and its results.
    \textbf{(b)} CIFAR-100-C corruption robustness experiment and its results. rmCE: relative mean corruption error w.r.t. LIF-based SNN (lower is better).
    \textbf{(c)} White-box adversarial robustness experiment on Fashion-MNIST and its results. rmAE: relative mean adversarial error w.r.t. LIF-based SNN (lower is better).
    \textbf{(d)} Black-box adversarial robustness experiment on Fashion-MNIST and its results.
    DendSNs in (c) and (d) use Stateful Dendrite and Mexican hat branch activation.
    \textbf{(e)} Mean squared distance between layer-2 somatic potentials under clean and noisy inputs.
    \textbf{(f)} t-SNE visualization of layer-2 somatic potentials under clean and noisy ($\epsilon = 0.5$) inputs.
    \textbf{(g)} Distributions of decision boundary thickness (Equation~\eqref{eq:methods-thickness}) across different models.
}
\label{fig:ml-robustness}
\end{figure}

Having demonstrated that dendritic computation can facilitate TIL when combined with task-specific modulation, we next examine its intrinsic benefits. In particular, we investigate whether DendSN alone can enhance the robustness of deep SNNs. Conventional ANNs and SNNs often struggle with corrupted inputs such as noisy data and adversarial attacks. In contrast, the human brain handles such challenges with ease. This performance gap partly stems from the fundamental differences between artificial and biological neural circuits. We contend that the integration of dendritic computing into neural networks has the potential to enhance robustness.

Noise robustness is crucial for deep learning models to maintain stable and reliable performance in real-world situations where data are susceptible to diverse sources of noise or corruption. To assess DendSNNs' robustness against noisy input, we conduct classification experiments on the Fashion-MNIST dataset \cite{xiao2017fashionmnist} with varying levels of Gaussian noise infused (Figure~\ref{fig:ml-robustness}(a), left). The models are first trained on clean training data and then evaluated on the test sets with different noise levels (detailed in Methods). As shown in Figure~\ref{fig:ml-robustness}(a), with the increase in noise level, the classification accuracies of all the models decrease. Nonetheless, DendSNNs consistently outperform LIF-based SNNs across all noise levels, showing better noise robustness.

However, Gaussian noise alone cannot cover the diverse corruption types in real life. To this end, we comprehensively test DendSNNs' robustness against various noise types on CIFAR-100-C \cite{hendrycks2019benchmarking}. The networks are first trained on the clean CIFAR-100 training set and frozen after that. Then, their error rates are obtained on the corrupted validation set, which comprises 19 corruption types, each with 5 levels of severity (Figure~\ref{fig:ml-robustness}(b), left). We aggregate the models' error rates across corruption types and severities using the relative mean corruption error (rmCE) metric \cite{hendrycks2019benchmarking} (Equation~\eqref{eq:methods-rmCE}, lower is better). On CIFAR-100-C, all DendSNNs yield significantly lower rmCEs than the baseline LIF-based SNN (the pink dashed line), indicating that DendSNNs are more resilient to corruptions than PointSNNs.

Besides noise, adversarial attacks represent another form of data corruption obtained by applying tiny yet intentionally worst-case perturbations to original samples \cite{goodfellow2014explaining}. Enhancing the robustness of neural networks against adversarial attacks is critical for security concerns. To check whether DendSNNs are less vulnerable to adversarial attacks than traditional SNNs, we conduct experiments based on the Fashion-MNIST dataset (Figure~\ref{fig:ml-robustness}(c), left). After training the models of interest on the original dataset, we employ the fast gradient sign method (FGSM) \cite{goodfellow2014explaining} to generate adversarial samples with respect to these models. The models' error rates on both the original and adversarial test sets under various perturbation amplitudes $\epsilon$ are recorded. Finally, we compute the relative mean adversarial error (rmAE), a metric similar to rmCE \cite{hendrycks2019benchmarking}, as a summarized metric (Equation~\eqref{eq:methods-rmAE}). As the accuracy curves show (Figure~\ref{fig:ml-robustness}(c), mid), DendSNN's classification accuracy decreases with a much slower rate than that of LIF-based SNN. All DendSNN conditions yield a significantly lower rmAE compared to the LIF-based SNN, and rmAE further decreases as the number of compartments $P$ grows (Figure~\ref{fig:ml-robustness}(c), right). These findings suggest that DendSNNs have a stronger resistance to adversarial attacks.

In the previous setup, adversarial attacks are applied directly to the models of interest. Consequently, the test set for one model is different from that of another, leading to unfairness. To make the comparisons more reasonable, we adopt a black-box adversarial attack setting (Figure~\ref{fig:ml-robustness}(d), left). This time, adversarial test samples are generated with regard to a shared baseline ANN. All the other settings are identical to the previous case. The final results exhibit a trend similar to the previous setting, vindicating the adversarial robustness of DendSNNs.

We further investigate the source of DendSNN's robustness gain using the noisy Fashion-MNIST task as an example. From the perspective of internal representations, Figure~\ref{fig:ml-robustness}(e) plots the mean squared distance between layer-2 somatic potentials under clean and noisy inputs (Equation~\eqref{eq:methods-potential-distance}). As noise intensity increases, this distance grows for all models, indicating feature drift due to input corruption. However, DendSNNs consistently exhibit smaller somatic potential distances than LIF-based SNNs, suggesting more stable internal representations. This stability is visually corroborated by the t-SNE \cite{maaten2008tsne} results in Figure~\ref{fig:ml-robustness}(f), where DendSNN features remain compact within each category even after noise injection, while those of LIF-based SNNs become dispersed. From the perspective of model predictions, we examine the distribution of decision boundary thickness (Equation~\eqref{eq:methods-thickness}) \cite{fawzi2018geometry,yang2020boundary}, which quantifies the magnitude of perturbation required to change a model's prediction. As shown in Figure~\ref{fig:ml-robustness}(g), DendSNNs exhibit higher mean thickness values, with fewer samples near the lower tail. Hence, DendSNNs predictions are less sensitive to input perturbations. Consistent with these observations, our earlier analysis of loss landscape (Figure~\ref{fig:dendsnn}(i)) shows that DendSNNs converge to flatter minima than LIF-based SNNs. Such a loss geometry with lower curvature indicates improved robustness \cite{moosavi2019curvature}. Taken together, dendritic computation promotes robustness by stabilizing internal representations, enlarging effective decision margins, and encouraging flatter loss landscapes.

\subsection*{Enhanced Few-Shot Learning Ability}
\label{subsec:results-fewshot}

\begin{figure}[t!]
\centering
\includegraphics[width=\linewidth]{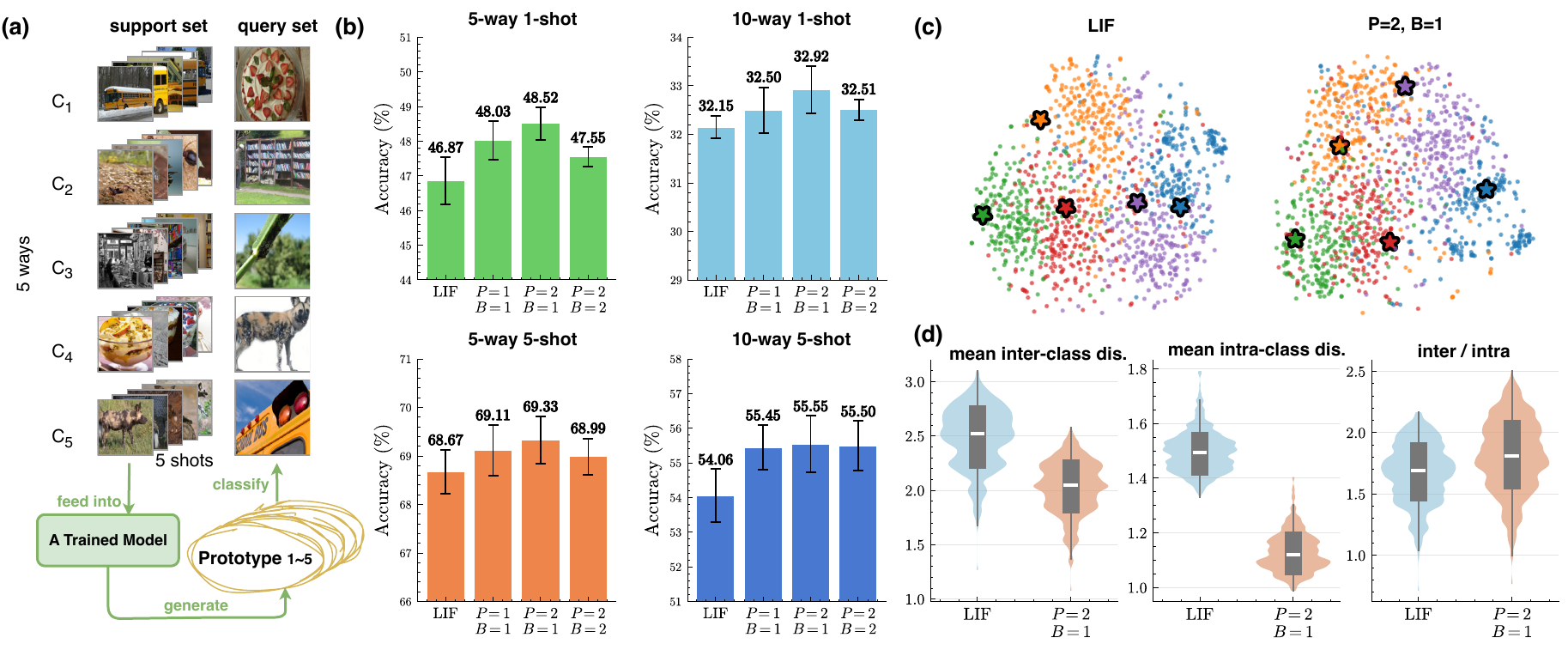}
\caption{
    \textbf{DendSNNs' few-shot learning performance.}
    \textbf{(a)} An illustration of few-shot learning's evaluation procedure and the Prototypical Network paradigm.
    \textbf{(b)} Different SNNs' few-shot learning performance on miniImageNet. 
    \textbf{(c)} t-SNE visualization of extracted features (dots) and prototypes (stars) for a 5-way 1-shot task. Colors indicate ground truth labels.
    \textbf{(d)} Distributions of mean inter-class distance (left), mean intra-class distance (mid), and the ratio between them (right). A larger ratio indicates better learned representations.
    DendSNs here use Stateless Dendrite and identity branch activation.
}
\label{fig:ml-fewshot}
\end{figure}

Conventionally, supervised training of deep neural networks relies heavily on a substantial amount of labeled data. Obtaining such data, however, is often costly and time-consuming. Consequently, it is crucial for machine learning models to generalize effectively to unseen domains and perform well when only limited labeled data are available. Few-shot learning methods aim to address this issue, but it remains challenging for conventional neural networks to extract transferable representations.

In the previous section, we showed that DendSNNs produce representations that are less sensitive to input perturbations, indicating that their learned features are more intrinsic and stable. Such stability suggests that DendSNNs may transfer more effectively to unseen samples. Moreover, biological observations indicates that dendrites support nonlinear integration and flexible representation of input patterns. Motivated by these insights, we hypothesize that incorporating dendritic computation into deep networks can strengthen the generalization of feature extractors, thereby enhancing few-shot learning performance.

We assess the few-shot learning capabilities of different backbones using the miniImageNet benchmark \cite{vinyals2017matching}. We adopt the Prototypical Network paradigm \cite{snell2017prototypical} (Figure~\ref{fig:ml-fewshot}(c)), and choose SEW ResNet-18 \cite{fang2021sew} as the backbone for feature extraction. A classical training pipeline is adopted, and the models are evaluated using 5-way 1-shot, 5-way 5-shot, 10-way 1-shot, and 10-way 5-shot classification accuracies. See Methods for more details about the training and evaluation protocols. As shown in Figure~\ref{fig:ml-fewshot}(b), all four types of accuracies significantly increase when the LIF-based backbone is replaced with a DendSNN, regardless of the value of $P$ and $B$. A t-SNE visualization \cite{maaten2008tsne} of an exemplary 5-way 1-shot task ($256$ query samples per class) shows that both the DendSNN and PointSNN can produce class-separable features (Figure~\ref{fig:ml-fewshot}(c)). Moreover, Figure~\ref{fig:ml-fewshot}(d) summarizes the feature distance statistics over $500$ randomly sampled 5-way 1-shot tasks (see Methods for details). DendSNNs exhibit smaller mean intra-class and inter-class distances than LIF-based SNNs; importantly, the reduction in mean intra-class distance is proportionally larger, resulting in higher inter-to-intra distance ratios. This pattern suggests that representations learned by DendSNNs are more compactly distributed in the embedding space while still preserving better inter-class separation, which makes the class prototypes more representative and discriminative. These findings highlight that DendSNNs are capable of extracting more generalizable features and distance metrics \cite{snell2017prototypical}, thereby offering a potential solution to alleviate the data-hungry bottleneck in various machine learning applications.

\section*{Discussion}
\label{sec:discussion}

Research in deep SNNs has advanced rapidly in recent years, with efforts spanning both network architectures and neuron modeling. The majority of existing deep SNNs are built upon point neuron models, which lack microscopic morphology and dendritic nonlinear dynamics that characterize real biological neurons. Consequently, the expressivity of these networks is limited by the simplicity of their underlying neurons \cite{poirazi2003pyramidal,jadi2014twolayer,tzilivaki2018dogma,beniaguev2021cortical}. Motivated by the insight that dendrites contribute substantially to biological neurons' representational capacity \cite{london2005dendritic,acharya2022dendritic}, we propose the dendritic spiking neuron (DendSN) model, aiming to empower deep SNNs with dendritic computation. Structurally, the model depicts dendritic morphology in two levels: compartments and branches. Computationally, hierarchical dendritic integration and nonlinearity are captured using a WNN-like formulation. DendSN exhibits significantly higher expressivity than point spiking neurons and can better approximate the biophysical activity pattern of a detailed multi-compartment neuron in the subthreshold regime. Moreover, networks of DendSNs trained through BPTT demonstrate superior supervised learning performance over SNNs based on point neurons, highlighting that enhanced single-neuron expressivity brought by dendritic computation can translate into improved network-level learning capacity.

The endeavor to introduce structural and dynamical complexity into neurons in deep SNNs, however, often brings two major challenges. First, increased internal complexity typically incurs excessive computational costs, limiting scalability to large models and complicated tasks. To reduce computational load, we assume that all compartments within a branch are arranged in parallel, thus preserving the two-level hierarchical structure while omitting inter-compartment connectivity. Passive compartmental dynamics are preserved, whereas active components are modeled only at branch junctions using $L_2$ aggregation and nonlinear activation. All DendSNs in a layer share the same dendritic morphology and branch configuration, which enables parallelism across compartments, branches, and neurons. Simplified variants such as Stateless Dendritic and Identity branch activation offer additional computational savings without performance tradeoff on certain tasks. We also implement Triton kernels for low-level acceleration. With all these designs, deep DendSNNs can be efficiently trained.

The second challenge is that morphologically detailed neuron models are less flexible, as their compatibility with diverse network structures is often constrained. Most existing dendritic neuron models are designed specifically for fully connected networks \cite{wu2018improved,yang2022biorealistic}, which prevents them from leveraging the advantages of modern deep networks. To ensure flexibility, DendSNs are designed to interface naturally with standard deep SNN components. Each DendSN layer accepts weighted features from preceding layers, which are partitioned along the channel dimension and mapped to dendritic compartments. Such a design makes DendSNs compatible with networks of arbitrary topology and depth.

Previous studies have demonstrated the crucial role of dendritic morphology and dynamics in neuronal information processing through detailed biophysical models of single neurons \cite{hay2011pyramidal,li2019dendritic,bicknell2021synaptic}. Reduced phenomenological models have also been employed to investigate how dendritic computation contributes to circuit-level computation and functions \cite{pagkalos2023dendrify}. However, in the context of deep neural networks, the impact of dendritic processing on network optimization and task performance remains largely unexplored. To disentangle this issue, we systematically evaluate DendSNNs across several challenging scenarios where conventional SNNs typically struggle. We first introduced dendritic branch gating (DBG), a biologically inspired algorithm that emulates dendritic modulation and synaptic clustering. DBG effectively mitigates catastrophic forgetting in task-incremental learning and can potentially be extended to other context-dependent learning paradigms such as multitask learning. Our experiments further revealed that DendSNNs exhibit enhanced robustness to noise and adversarial attacks, as well as improved generalization in few-shot learning tasks. Collectively, these findings highlight the advantages of integrating dendritic computation into deep SNNs, suggesting that DendSNNs may offer a more biologically grounded and resilient approach for real-world neural computation.

In future work, we plan to extend DendSNNs to deeper architectures and larger-scale tasks (\eg ImageNet classification \cite{deng2009imagenet}) to further narrow the gap between DendSNNs and state-of-the-art deep network models. This will require both refined architectural designs and dedicated training methodologies. We will also explore the broader applicability of DendSNNs beyond visual classification, particularly in domains such as brain-machine interfaces, where low energy consumption, robustness, and representational capacity are essential. Furthermore, we intend to enhance the biological plausibility of DendSNNs by designing biologically inspired learning rules that exploit dendritic states for effective deep network training. This effort aligns with our goal of advancing DendSNNs beyond traditional simulations, making DendSNNs valuable tools for designing and implementing brain-inspired learning rules on deep networks. A detailed discussion of the current limitations and potential research directions can be found in Supplementary Information \ref{supsec:limitation}.

In summary, our work opens a promising avenue for integrating dendritic computing into deep SNNs for practical machine learning applications, broadening the impact of dendrite modeling beyond neuroscience simulations.
\section*{Methods}
\label{sec:methods}

\subsection*{Dendritic spiking neuron}
\label{subsec:methods-dendsn}

The dendritic spiking neuron (DendSN) model simplifies dendritic morphology by neglecting the fine-grained connectivity among compartments within individual branches. It considers only which dendritic branch each compartment belongs to. To be specific, a DendSN comprises $P$ compartments located on $B$ segregated dendritic branches and one soma. To describe the dendritic morphology, the dendritic wiring matrix $\Gamma \in \{0,1\}^{B\times P}$ is defined such that $\Gamma_{b,i} = 1$ if compartment $i$ is on branch $b$, and $\Gamma_{b,i} = 0$ otherwise. Also, denote the set of compartments on branch $b$ as $C_b = \{i \mid \Gamma_{b,i}=1\}$. Since each compartment $i$ is assigned to only one of the $B$ branches, we have $\cup_{b=1}^B C_b = \{1, \dots, P\}$ and $C_b \cap C_{b'} = \emptyset \ (b \ne b')$. All $B$ branches are connected to the soma directly.

The dynamics of the $i$-th dendritic compartment can be described as
\begin{equation}
\label{eq:methods-compartment-continuous}
\tau_d \frac{\diff v^d_i}{\diff t} = - (v_i^d - v^d_{\mathrm{rest}}) + x_i(t),
\end{equation}
where $v^d_i$ is the local potential, $\tau_d$ is the time constant, $v^d_{\mathrm{rest}}$ is the dendritic resting potential, and $x_i(t)$ is the synaptic input. For efficient implementation, we assume $v_{\mathrm{rest}}^d=0$ and discretize Equation \eqref{eq:methods-compartment-continuous} into $T$ time steps \cite{wu2019direct,fang2023spikingjelly} using the Euler method
\begin{equation}
\label{eq:methods-compartment}
V^d_i[t] = \alpha V_i^d[t-1] + X_i[t].
\end{equation}
Here, $t \in \{1, \dots, T\}$ is the time step index, $0 \le \alpha < 1$ is the dendritic decay factor, while $V_i^d$ and $X_i$ correspond to $v^d_i$ and $x_i$ in Equation \eqref{eq:methods-compartment-continuous}, respectively \footnote{Throughout this work, uppercase letters represent variables in discrete time steps, while corresponding lowercase letters are used for continuous time variables. Unless otherwise specified, variables in regular font are scalars, while those in bold font are vectors, matrices, or tensors.}. $V^d_i$ is initialized to the resting state, \ie $V^d_i[0] = 0$. Notice that $\tau_d$ and the size of time step are absorbed into $\alpha$ and $X_i[t]$ for simplicity.

The input signal from branch $b$ to the soma at time $t$, represented as $Y_b[t]$, is determined by the instantaneous local potentials of all compartments on branch $b$. This aggregation process is defined as
\begin{equation}
\label{eq:methods-branch-aggregation}
Y_b[t] = \psi\left\{ \left[ \sum_{i \in C_b} \left(\frac{V_i^d[t] - \xi_i}{\zeta_b}\right)^2\right]^{\frac{1}{2}} \right\},
\end{equation}
where $\xi_i \in \mathbb{R}$ is the translation factor for compartment $i$, $\zeta_b \in \mathbb{R}^+$ is the dilation factor for branch $b$, and $\psi$ is the nonlinear dendritic activation function. Inspired by theories of wavelet analysis and wavelet neural networks (WNNs) \cite{zhang1995wnn,alexandridis2013wnnguide,liu2021wcnn}, we use the Mexican hat wavelet as $\psi$ to enhance the expressivity of DendSN:
\begin{equation}
\label{eq:methods-mexican-hat}
\psi(x) = (1-x^2)e^{-\frac{1}{2}x^2}.
\end{equation}
Alternatively, we can use identity mapping $\psi(x) = x$ to simplify the model when dendritic nonlinearity is not necessary. See Figure~\ref{supfig:psi} for an illustration of different $\psi$. The total input signal to the soma at time $t$ is the weighted sum of $Y_b[t]$:
\begin{equation}
\label{eq:methods-soma-input}
Z[t] = \sum_{b=1}^B \kappa_b Y_b[t],
\end{equation}
where $\kappa_b \in \mathbb{R}$ is the weight for branch $b$. $\kappa_b$ is also referred to as branch strength.

The soma can be any point spiking neuron model. Take the leaky integrate-and-fire (LIF) neuron as an example \cite{lapicque1907lif,gerstner2014neuronal},
\begin{align}
    V^s[t] &= \beta (1-S[t-1])V^s[t-1] + Z[t], \label{eq:methods-lif-reset-charge} \\
    S[t] &= \Theta(V^s[t] - 1), \label{eq:methods-lif-fire}
\end{align}
where $V^s$ is the somatic membrane potential, $0 \le \beta < 1$ is the decay factor of $V^s$, $\Theta(x)$ is the Heaviside step function (yields $1$ if $x\ge 0$ and $0$ otherwise), and $S[t] \in \{0, 1\}$ is the binary output signal at time step $t$ ($1$ means firing and $0$ means not firing). Here we assume that the somatic resting and reset potentials are both $0$, and the firing threshold is set to $1$. The reset process is absorbed into Equation~\eqref{eq:methods-lif-reset-charge} for brevity. The initial condition is $V^s[0]=0$. Unless otherwise specified, DendSNs in this work use LIF as the soma model. Other somatic models discussed in this work include parametric LIF (PLIF) \cite{fang2020plif} and sliding parallel spiking neuron (SPSN) \cite{fang2023psn}.

Viewing the neuron as a whole, an individual DendSN maps synaptic inputs $\mathbf{X}$ to somatic spikes $\mathbf{S}$. The DendSN model can be summarized as
\begin{equation}
\label{eq:dendsn}
\begin{aligned}
    \mathbf{X} &\triangleq \{X_i[t] \mid t\in \{1, \dots, T\}, i \in \{1, \dots, P\}\},\ \mathbf{S} \triangleq \{S[t] \mid t\in \{1, \dots, T\}\}, \\
    \mathbf{S} &= \mathrm{DendSN}(\mathbf{X}; \alpha, \psi, \Gamma, \{\xi_i\}_{i=1}^P, \{\zeta_b\}_{b=1}^B, \{\kappa_b\}_{b=1}^{B}, \dots),
\end{aligned}
\end{equation}
where somatic parameters (\eg $\beta$ in Equation~\eqref{eq:methods-lif-reset-charge}) depend on the specific soma type and thus are omitted here. $\alpha$, $\{\xi_i\}_{i=1}^P$, $\{\zeta_b\}_{b=1}^B$, and $\{\kappa_b\}_{b=1}^{B}$ are learnable.

\subsection*{Details of L5PC approximation}
\label{subsec:methods-l5pc}

Beniaguev et al. constructed a multi-compartment biophysical model of a layer-5 pyramidal cell (L5PC) from a rat's brain and collected its activity data \cite{beniaguev2021cortical}. We aim to predict the somatic membrane potential of the fine-grained model given the presynaptic spikes using a reduced neuron model. First, the simulation data are binned into $6000$ time steps, with $1$ms per step. Each sample is then cropped to the first $T=2048$ time steps. A total of $N=1278$ channels of presynaptic spikes are recorded (Figure~\ref{fig:dendsn}(b)), with each synaptic site contributing an excitatory and an inhibitory channel ($1278/2=639$ synaptic sites). Among these, $377$ synaptic sites are on the apical dendrite, producing $N_{\mathrm{a}}=754$ apical channels. The remaining $N_{\mathrm{b}}=524$ channels come from $262$ synaptic sites on the basal dendrite. The dataset is split into training, validation, and test sets in an $8:1:1$ ratio.

For DendSNs in this experiment, there are $B=2$ branches representing the apical and basal dendrites. Each branch has two compartments ($P=2\times 2 = 4$), one receiving inputs from the excitatory channels on the branch and the other receiving inhibitory inputs. The input spikes from $N=1278$ channels are weighted by synaptic strengths $\{W_j\}_{j=1}^{N}$, fed into their target compartments, and summed locally (see the red shaded areas in Figure~\ref{fig:dendsn}(b)). For point neurons including LIF, PLIF, and SPSN, spikes are weighted, summed, and directly fed into the soma at each time step. The resulting somatic potential is taken as the predicted potential $\widehat{V^s}[t]$. The loss combines a mean squared error (MSE) term for somatic potential regression and a binary focal loss \cite{lin2017focal} for spike prediction:
\begin{equation}
\label{eq:methods-l5pc-loss}
\mathcal{L} = \frac{1}{T}\sum_{t=1}^{T}\left(V^s[t] - \widehat{V^s}[t]\right)^2
+ \frac{\lambda}{T}\sum_{t=1}^{T} \left[S[t]\alpha^{+} + (1 - S[t])\alpha^{-}\right]\left\{1 - \mathrm{exp}\left[-\mathrm{BCE}(\widehat{P}[t], S[t])\right]\right\}^\gamma \mathrm{BCE}(\widehat{P}[t], S[t]),
\end{equation}
where $V^s[t]$ is the ground truth somatic potential, $S[t]\in \{0,1\}$ is the ground truth spike signal, $\alpha^{+}=0.9$ and $\alpha^{-}=0.1$ are the weights for spike and non-spike cases, $\gamma=2$ is the focusing parameter, $\mathrm{BCE}$ denotes the binary cross-entropy loss, and $\lambda=2000$ controls the weight of the focal loss term. $\widehat{P}[t]$ is the predicted spike probability defined as
\begin{equation}
\label{eq:methods-l5pc-spike-prob}
\widehat{P}[t] = \sigma\left(4\cdot \frac{\widehat{V^s}[t]-V_{\mathrm{th}}}{V_{\mathrm{th}}-V_{\mathrm{reset}}}\right),
\end{equation}
where $\sigma$ is the sigmoid function, and $V_{\mathrm{th}}$ and $V_{\mathrm{reset}}$ are the firing threshold and reset potentials derived from the L5PC model. Both synaptic weights and learnable neuronal parameters are optimized via backpropagation through time (BPTT) and surrogate gradient (see Equation~\eqref{eq:methods-sg}). Other hyperparameters are listed in Table~\ref{suptab:hyperparameters}.

We evaluate prediction quality on the test set using the coefficient of determination ($R^2$) between the predicted somatic potential sequence $\{\widehat{V}[t]\}_{t=1}^{T}$ and ground truth $\{V[t]\}_{t=1}^{T}$. For a test set of $M_{\mathrm{test}}$ samples:
\begin{equation}
\label{eq:methods-l5pc-r2}
R^2 = 1 - \frac{\sum_{m=1}^{M_{\mathrm{test}}}\sum_{t=1}^{T} (V_m[t] - \widehat{V}_m[t])^2}{\sum_{m=1}^{M_{\mathrm{test}}}\sum_{t=1}^{T} (V_m[t] - \bar{V})^2},
\end{equation}
where $V_m[t]$ and $\widehat{V}_m[t]$ are the ground-truth and predicted potentials for sample $m$ at time step $t$, and $\bar{V} = \frac{1}{M_{\mathrm{test}}T}\sum_{m=1}^{M_{\mathrm{test}}}\sum_{t=1}^{T} V_m[t]$ is the mean somatic potential across all test samples and time steps.

\subsection*{DendSN variants}
\label{subsec:methods-variants}

The dendrite model defined in Equations~\eqref{eq:methods-compartment} to \eqref{eq:methods-soma-input} serves as the default configuration of DendSN, unless otherwise specified. We further investigate several model variants to analyze the functional role of dendritic components or to boost computational efficiency and task performance. For each experiment, we evaluate all available variants of DendSNN and report the results obtained from the best-performing variant.

\paragraph{Activation-free branch}
By default, the dendritic activation function is the Mexican hat wavelet (Equation~\eqref{eq:methods-mexican-hat}). As an alternative, we remove the nonlinear activation by using an identity mapping $\psi(x)=x$. The variant is named \textbf{Identity}, while the default is denoted as \textbf{MH}. The Identity variant is used (1) to evaluate the contribution of dendritic nonlinearity in single-neuron simulation tasks and (2) to alleviate the problem of gradient vanishing in deep networks, since the magnitude of the standard Mexican hat function's derivative is smaller than $1$ for most input values (see Figure~\ref{supfig:psi}). Note that there is no significant difference between the computational cost of these two variants (Figure~\ref{fig:dendsnn}(j)).

\paragraph{Product-based aggregation}
In the default model, dendritic branch $b$ aggregates compartment signals by $L_2$-norm (Equation~\eqref{eq:methods-branch-aggregation}). We also consider a product-based alternative:
\begin{equation}
\label{eq:methods-branch-aggregation-product}
Y_b[t] = \psi\left( \prod_{i \in C_b} \frac{V_i^d[t] - \xi_i}{\zeta_i} \right).
\end{equation}
Here, the dilation parameter $\zeta_i$ is assigned per compartment, following common WNN practice \cite{alexandridis2013wnnguide}. This variant is used to study how different aggregation rules affect single-neuron behavior. Due to its susceptibility to gradient vanishing, it is not adopted in deep learning experiments.

\paragraph{Stateless compartment}
By default, each dendritic compartment acts as a leaky integrator (Equation~\eqref{eq:methods-compartment}). Alternatively, we remove compartmental dynamics by setting $V_i^d[t] = X_i[t]$. This configuration eliminates the need to maintain compartmental potentials (\aka states), so it is referred to as \textbf{Stateless Dendrite}; the default condition is called \textbf{Stateful Dendrite}. Due to the higher computational efficiency (Figure~\ref{fig:dendsnn}(j)), Stateless Dendrite is widely used in deep learning experiments where temporal dependencies are less critical.

\paragraph{Residual dendrite}
In the default model, the somatic input $Z[t]$ is a weighted sum of dendritic branch outputs $Y_b[t]$ (Equation~\eqref{eq:methods-soma-input}). Optionally, we can introduce a residual connection from the synaptic inputs $\{X_i[t]\}_{i=1}^{P}$ so that
\begin{equation}
\label{eq:methods-residual}
Z[t] = \sum_{b=1}^B \kappa_b Y_b[t] + \frac{1}{P}\sum_{i=1}^P X_i[t].
\end{equation}
Notice that $\{X_i[t]\}_{i=1}^{P}$ is averaged before being added to $Z[t]$ to ensure shape compatibility. We refer to Stateful (Stateless) Dendrite with a residual connection as \textbf{ResStateful} (\textbf{ResStateless}) \textbf{Dendrite}. The residual connection facilitates training of deep DendSNNs on complex tasks.

\subsection*{Dendritic spiking neural networks}
\label{subsec:methods-dendsnn}

To integrate DendSNs into deep SNN architectures, they are organized into layers. Analogous to point spiking neuron layers in conventional SNNs, DendSN layers are positioned after weight layers, acting as activation functions. For simplicity and computational efficiency, the following constraints are imposed on DendSN layers:
(1) Each dendritic branch contains an equal number of compartments ($P/B$).
(2) All DendSNs within a layer share the same number of branches $B$ and compartments $P$.
(3) Neurons on different channels have distinct sets of learnable neuronal parameters, whereas neurons at different spatial positions within the same channel share a common parameter set; an exception is the compartmental decay factor $\alpha$, which is shared by all neurons in the layer, following the practice of PLIF \cite{fang2020plif}.

As indicated by the shapes of $\mathbf{X}$ and $\mathbf{S}$ in Equation~\eqref{eq:dendsn}, a DendSN layer reduces the feature size by a factor of $P$, differing from point neuron layers that preserve the original tensor shape. This compression is performed along the channel dimension to fuse information from different feature maps at each spatial location. Formally, let $\tilde{\mathbf{X}}[t] \in \mathbb{R}^{C \times \dots}$ denote the output of the previous weight layer at time step $t$, where $C = C_o P$ and the ellipsis indicate spatial dimensions (e.g., $\tilde{\mathbf{X}}[t] \in \mathbb{R}^{C}$ for fully connected layers, or $\tilde{\mathbf{X}}[t] \in \mathbb{R}^{C \times H \times W}$ for 2D convolutional layers). The feature is reshaped into $\mathbf{X}[t] \in \mathbb{R}^{C_o \times P \times \dots}$, where the first dimension indexes DendSN layer's output channels, and the second dimension indexes dendritic compartments. Each group of $P$ consecutive feature maps from $\tilde{\mathbf{X}}[t]$ thus serves as the inputs to a neuron channel composed of DendSNs with $P$ compartments. Consequently, the DendSN layer outputs a spike tensor of shape $(C_o, \dots)$ at each time step.

By stacking multiple weight-DendSN blocks, deep dendritic spiking neural networks (DendSNNs) of arbitrary architecture can be constructed. A DendSNN architecture can also be derived from an existing PointSNN by replacing neurons and adjusting the number of channels. Specifically, consider two consecutive weight-neuron blocks in a PointSNN:
\begin{equation}
\label{eq:methods-dendsnn-from}
\text{Proj}(\text{in}=C_1, \text{out}=C_2) \to \text{PointNeuron}(\text{out}=C_2) \to \text{Proj}(\text{in}=C_2, \text{out}=C_3) \to \text{PointNeuron}(\text{out}=C_3),
\end{equation}
where \text{Proj} denotes a weight (projection) layer. The subnetwork can be replaced with
\begin{equation}
\label{eq:methods-dendsnn-to}
\text{Proj}(\text{in}=C_1, \text{out}=C_2) \to \text{DendSN}(\text{out}=C_2/P) \to \text{Proj}(\text{in}=C_2/P, \text{out}=C_3 P) \to \text{DendSN}(\text{out}=C_3).
\end{equation}
Notice that the parameter count is dominated by the weights of the projection layers. The first projection layer maintains the same weight dimensions, while the second projection layer, after scaling the dimensions, also has an unchanged number of elements in the weight matrix ($C_2\times C_3$). Consequently, converting a PointSNN into a DendSNN does not significantly increase the parameter count. See Supplementary Materials \ref{supsec:parameter-counts} for a detailed analysis.

The dynamics of a DendSN layer are fully differentiable, except for the Heaviside step function $\Theta$ in the somatic spike generation process (\eg Equation~\eqref{eq:methods-lif-fire}). This issue can be circumvented by surrogate gradient \cite{neftci2019surrogate,zenke2021remarkable}, \ie using the derivative of a smooth surrogate function to approximate the derivative of $\Theta$. In this work, we adopt the arctangent surrogate function:
\begin{equation}
\label{eq:methods-sg}
\Theta'(x) \approx \frac{\diff}{\diff x}\left[\frac{1}{\pi}\arctan(\pi x) + \frac{1}{2}\right] = \frac{1}{1 + \pi^2 x^2}
\end{equation}
With Equation~\eqref{eq:methods-sg}, DendSNNs can be trained directly in a supervised manner using backpropagation through time (BPTT) \cite{wu2018stbp,zenke2018superspike,shrestha2018slayer}.

\subsection*{Details of classification experiments}
\label{subsec:methods-classification}

Three classification tasks were conducted to evaluate the proposed DendSNNs, including Fashion-MNIST~\cite{xiao2017fashionmnist}, CIFAR10-DVS~\cite{li2017cifar10dvs}, and Tiny ImageNet. These tasks are designed to progressively test the scalability of DendSNNs under increasing network depth and task complexity. For all three tasks, the input data is directly fed into the model so that the first weight-neuron block serves as a learnable spike encoder \cite{rathi2023diet}. All models are trained using BPTT and surrogate gradient \cite{wu2018stbp,zenke2018superspike,shrestha2018slayer} (Equation~\eqref{eq:methods-sg}).

\paragraph{Fashion-MNIST}
Fashion-MNIST \cite{xiao2017fashionmnist} is a grayscale image dataset containing $70,000$ samples ($60,000$ training samples and $10{,}000$ test samples) of $28\times28$ resolution, categorized into 10 classes. It is considered a harder version of MNIST \cite{lecun1998mnist}. No data augmentation is applied. Fully connected SNNs with two hidden layers are trained; the first hidden layer contains $2000/P$ neurons, and the second hidden layer contains $2000$ neurons. Hyperparameters are provided in Table~\ref{suptab:hyperparameters}.

\paragraph{CIFAR10-DVS}
CIFAR10-DVS \cite{li2017cifar10dvs} is a neuromorphic vision dataset converted from the standard CIFAR-10 images \cite{krizhevsky2009cifar} using a Dynamic Vision Sensor \cite{lichtsteiner2008dvs}. It contains $10,000$ event streams from 10 categories with 2 channels and $128\times 128$ resolution. Following the practice of previous works \cite{duan2022tebn,fang2023psn}, we split the dataset into $9{,}000$ training samples and $1{,}000$ test samples. The spatial resolution is downsampled to $32\times 32$, and each sample is integrated into $T=10$ frames. Data augmentation process involves random resized cropping, random horizontal flipping, and Neuromorphic Data Augmentation (NDA) \cite{li2022nda}. The network architecture is adapted from Spiking VGG-11 \cite{duan2022tebn,fang2023psn}; for DendSNNs, the first two layers remain point-neuron based rather than being replaced with DendSNs, which empirically leads to higher final accuracy. For other hyperparameters, refer to Table~\ref{suptab:hyperparameters}.

\paragraph{Tiny ImageNet}
Tiny ImageNet is a subset of ImageNet-1k \cite{deng2009imagenet} containing $200$ classes, each with $500$ training images and $50$ validation images, totaling $100,000$ training and $10,000$ validation samples. Each image has three color channels and is cropped to a spatial resolution of $64\times64$. The data augmentation pipeline incorporates random horizontal flipping, AutoAugment \cite{cubuk2019autoaugment} and normalization. The model architecture is based on Spiking VGG-13 \cite{huang2024clif}, with the classification head simplified to a single fully connected layer; DendSN replacement is applied only to the last four convolution-neuron blocks, which empirically leads to a better performance. Other hyperparameters are listed in Table~\ref{suptab:hyperparameters}.

\subsection*{Dendritic branch gating for task incremental learning}
\label{subsec:methods-dbg}

Inspired by dendritic modulation \cite{wybo2023NMDA} and synaptic clustering \cite{cichon2015branch,limbacher2020cluster} in neuroscience, we design a new algorithm named dendritic branch gating (DBG) for task incremental learning (TIL) using DendSNns. The main idea is to modulate the dendritic branch strengths of all the DendSNs in a network using the index of the current task.

Suppose a DendSNN is going to sequentially learn $Q$ tasks indexed $q\in \{1, \dots, Q\}$. The task index $q$ is fed into the network as a top-down modulation input together with the bottom-up feedforward input during training and inference, which is a common practice in previous works like XdG \cite{masse2018xdg}. For each DendSN layer in the DendSNN, DBG modulates the dendritic branch strength vector by applying a task-specific binary mask
\begin{equation}
\label{eq:methods-dbg-mask}
    \mathbf{K}^{*} = \mathbf{K} \odot \mathrm{DBG}(q;\mathbf{K}, \rho),
\end{equation}
where $\mathbf{K}=\left[\kappa_{i,b}\right]_{C\times B}$ is the unmasked branch strength matrix of a DendSN layer ($C$ is the number of DendSNs along the channel dimension, and $\kappa_{i,b}$ is the strength of the $b$-th branch for DendSNs at the $i$-th channel), $\mathrm{DBG}(q;\mathbf{K}, \rho) \in \{0,1\}^{C\times B}$ is a sparse binary mask with the same shape as $\mathbf{K}$, $\rho$ controls sparsity, and $\mathbf{K}^{*}$ is the masked strength matrix. Each element $m_{i,b}$ of the mask is independently sampled from a Bernoulli distribution with parameter $\rho$
\begin{equation}
\label{eq:methods-dbg-sampling}
\mathrm{DBG}(q;\mathbf{K}, \rho) = \left[m_{i,b}\right]_{C\times B}, \ \text{where } m_{i,b} \sim \mathrm{Bernoulli}(\rho),\ \forall i,b.
\end{equation}
The value of $\mathrm{DBG}(q;\mathbf{K}, \rho)$ is fixed after sampling (\ie once $q$, $\rho$, and $\mathbf{K}$ are the same, $\mathrm{DBG}$ deterministically returns the same mask), while the dense matrix $\mathbf{K}$ remains learnable. The masking procedure is applied during both training and inference. See Supplementary Materials \ref{supsec:dbg-pseudocode} for pseudocode.

To investigate the impact of sparsity, we train DendSNNs with different $\rho$ values. We also propose a variant of DBG called DBG-embedding ($\mathrm{DBG}_e$), which directly produces a dense but learnable branch strength matrix for each task $q$ and each layer
\begin{equation}
\label{eq:methods-dbg-embedding}
    \mathbf{K}^{*} = \mathrm{DBG}_e(q;\mathbf{K}),
\end{equation}
where $\mathrm{DBG}_e(q;\mathbf{K}) \in \mathbb{R}^{C\times B}$ is task-specific. For all $q\in \{1,\dots,Q\}$, the matrix $\mathrm{DBG}_e(q;\mathbf{K})$ is learnable and is optimized together with other model parameters. The difference between DBG with $\rho=1$ and $\mathrm{DBG}_e$ is that the former shares the same unmasked strength matrix $\mathbf{K}$ across all tasks, while the latter maintains separate strength matrices for different tasks.

\subsection*{Details of task incremental learning experiments}
\label{subsec:methods-til}

We denote a task incremental learning scenario as $\{\mathcal{I}_q\}_{q=1}^{Q}$, where $\mathcal{I}_q$ is the $q$-th task.
A model should be trained on these tasks sequentially, using a supervised learning setting similar to that described in Subsection \textit{Details of classification experiments}. When training on task $q$, the model has no access to data on tasks $1$ to $q-1$. To evaluate the models' resistance to catastrophic forgetting, we use the average accuracy as the metric
\begin{equation}
\label{eq:methods-til-avg-acc}
    \overline{acc}_q = \frac{1}{q}\sum_{j=1}^{q} acc_{q,j}, \ \forall q \in \{1, \dots, Q\}
\end{equation}
where $acc_{q,j}$ is the test accuracy on task $j$ ($1\le j \le q$) after training the model sequentially on task $1$ to $q$.

The Permuted MNIST \cite{goodfellow2013empirical,masse2018xdg} experiment consists of $Q=50$ image classification tasks, each of which is a different pixel-level permutation of the entire original MNIST task \cite{lecun1998mnist} (\ie all the samples in the original dataset are permuted using $\mathrm{Perm}_q$). See Figure~\ref{fig:dbg}(d) for an illustration. We adopt $50$ random seeds to generate these permutations, and all runs share the same set of seeds. Network architecture and hyperparameter settings are the same as those in the Fashion-MNIST experiment (Subsection \textit{Details of classification experiments}).

The proposed DBG algorithm is an architecture-based TIL method \cite{wang2023continual}. To further enhance TIL performance, we additionally employ elastic weight consolidation (EWC) \cite{kirkpatrick2017overcoming}, a regularization-based approach inspired by synaptic consolidation, either on the decoder layer only (dEWC) or to the full network (fEWC). Combining DBG and EWC is biologically plausible, as synaptic clustering (DBG) and synaptic consolidation (EWC) coexist in neural circuits and complementarily contribute to continual learning. The EWC regularization coefficient is set to $20,000$, selected through hyperparameter search.

To analyze why DendSNN and DBG improve TIL performance, we measure the mean squared synaptic distance between the weights after learning earlier tasks and those after completing all tasks. Let $\mathbf{W}^{l,q} \in \mathbb{R}^{M\times N}$ denote the weight matrix of layer $l$ after training on the first $q$ tasks, and $\mathbf{W}^{l,Q}$ denote the final weight matrix after training on all $Q$ tasks. The mean squared synaptic distance is defined as
\begin{equation}
\label{eq:methods-synaptic-distance}
d^l_q =
\begin{cases}
\displaystyle
\frac{1}{MN}\|\mathbf{W}^{l,Q} - \mathbf{W}^{l,q}\|^2_F & \text{without DBG},\\[8pt]
\displaystyle
\frac{1}{|\Omega^{l,1}|}\left\|\mathbf{M}^{l,q} \odot (\mathbf{W}^{l,Q} - \mathbf{W}^{l,q})\right\|^2_F & \text{with DBG},
\end{cases}
\end{equation}
where $\|\dots\|_F$ is the Frobenius norm, $\mathbf{M}^{l,q} \in \{0,1\}^{M\times N}$ is a binary mask indicating the active synaptic connections for task $q$ when DBG is applied, $\odot$ denotes element-wise multiplication, and $|\Omega^{l,q}|$ is the number of active synapses in $\mathbf{W}^{l,q}$. In other words, it is the mean squared error between the activated synaptic weights. A larger $d^l_q$ indicates greater deviation of synaptic weights, implying stronger interference and weaker knowledge retention.

\subsection*{Details of noise robustness experiments}
\label{subsec:methods-noise-robustness}

In the Fashion-MNIST noise robustness experiment, we add Gaussian noise with amplitude $0, 0.05, 0.10, \dots, 0.50$ on the test set of Fashion-MNIST ($11$ noise levels). A model is first trained on the clean training set, and then evaluated on test sets with all noise levels. The network architecture and hyperparameter settings follow those used for Fashion-MNIST classification (Subsection \textit{Details of classification experiments}). Accuracies on noisy test sets are used for evaluation.

CIFAR-100-C \cite{hendrycks2019benchmarking} is a corruption robustness benchmark based on CIFAR-100 \cite{krizhevsky2009cifar} containing test samples with $N_c=19$ types of corruptions, each with $N_s=5$ levels of severity (Figure~\ref{fig:ml-robustness}(b), left). Each sample is an RGB image of size $32\times 32$, and there are $100$ classes. We first train a MS-ResNet18 \cite{hu2024msresnet} (with LIF or DendSN) on the clean CIFAR-100 training set, where random cropping, random horizontal flipping, AutoAugment \cite{cubuk2019autoaugment}, Cutout \cite{devries2017cutout} and normalization are applied to augment training data. DendSN replacement is not applied to the first two residual blocks, which empirically leads to better results. The trained model is then frozen and evaluated on the $N_c\times N_s$ corrupted test sets. We use the relative mean corruption error (rmCE) \cite{hendrycks2019benchmarking} as the evaluation metric
\begin{equation}
\label{eq:methods-rmCE}
\mathrm{rmCE}^f = \frac{1}{N_c} \sum_{c=1}^{N_c} \left[ \frac{\sum_{s=1}^{N_s} (\mathrm{acc}_{\mathrm{clean}}^f - \mathrm{acc}_{s,c}^f)}{\sum_{s=1}^{N_s} (\mathrm{acc}_{\mathrm{clean}}^{f_b} - \mathrm{acc}_{s,c}^{f_b})} \right],
\end{equation}
where $f$ is the model of interest, $\mathrm{acc}_{s,c}^f$ is the accuracy of $f$ on the test set with corruption type $c$ and severity $s$, $\mathrm{acc}_{\mathrm{clean}}^f$ is the accuracy of $f$ on the clean test set, and $f_b$ refers to baseline model (\ie LIF-based SNN). For hyperparameters, see Table \ref{suptab:hyperparameters}.

To examine how DendSN enhances network robustness against input noise, we quantify the deviation of neuronal responses by measuring the mean squared distance between somatic potentials obtained from clean and noisy inputs. Let $\textbf{V}^{s,l}_{\epsilon} \in \mathbb{R}^{T \times D}$ denote the sequence of somatic potentials in layer $l$ under a noise level $\epsilon$, where $T$ and $D$ are the number of time steps and neurons. The mean squared potential distance is defined as
\begin{equation}
\label{eq:methods-potential-distance}
d^l_\epsilon = \frac{1}{TD} \|\textbf{V}^{s,l}_{\text{clean}} - \textbf{V}^{s,l}_{\epsilon}\|_F^2 ,
\end{equation}
where $\textbf{V}^{s,l}_{\text{clean}}$ represents the corresponding potentials under clean inputs. In other words, $d^l_\epsilon$ is the mean squared error between the somatic potential sequences under clean and noisy inputs. Smaller values of $d^l_\epsilon$ indicate that the internal representations are more stable and thus more resilient to noise corruption.

Additionally, we report the distribution of decision boundary thickness for each trained model, which characterizes how gradually the model's predictions change in the input space and serves as an indicator of noise robustness. Formally, for a given input $\mathbf{x} \in \mathbb{R}^{D_{\text{in}}}$, small perturbations are added as $\mathbf{x}'= \mathbf{x} + \delta\mathbf{r}$, where $\mathbf{r} \in \mathbb{R}^{D_{\text{in}}}$ is a random unit vector and $\delta$ controls the perturbation magnitude. The margin $\Delta$ for $\mathbf{x}$ along a direction $\mathbf{r}$ is defined as the minimum $\delta$ that changes the predicted class, and the decision boundary thickness $\mathcal{T}$ is the minimum margin across all directions:
\begin{align}
\Delta(\mathbf{x} ; \mathbf{r}, f) &= \min \left\{ \delta \mid f(\mathbf{x}+\delta\mathbf{r}) \ne f(\mathbf{x}) \right\}, \label{eq:methods-margin} \\
\mathcal{T}(\mathbf{x} ; f) &= \min \left\{\Delta(\mathbf{x} ; \mathbf{r}, f) \mid \forall \mathbf{r} \right\}. \label{eq:methods-thickness}
\end{align}
Here, $f(\mathbf{x})$ denotes the trained network's category prediction. Since computing $\mathcal{T}$ exactly is intractable, we approximate it using Monte Carlo sampling. For each input $\mathbf{x}$, 250 random unit vectors $\mathbf{r}$ are sampled; for each direction, $\delta \in [0, 10]$ is searched with step size $0.02$ to find the smallest perturbation that changes the predicted class. The estimated thickness is taken as the minimum margin across all sampled directions. We evaluate this metric on 128 randomly selected clean test samples shared across all models, yielding 128 thickness measurements per model. The resulting distributions are visualized in Figure~\ref{fig:ml-robustness}(f). Larger thickness values indicate a more stable decision surface, suggesting that the model is more robust to input perturbations.

\subsection*{Details of adversarial robustness experiments}
\label{subsec:methods-adversarial-robustness}

In Fashion-MNIST adversarial robustness experiments, models of interest are first optimized on the clean Fashion-MNIST training set and evaluated on the clean test set. The network architecture and training hyperparameters are identical to those used for Fashion-MNIST classification (Subsection \textit{Details of classification experiments}). Then, adversarial test samples are generated. For the ``attack-on-model'' (\aka ``white-box attack'') case, adversarial samples with respect to the models of interest are directly generated using the fast gradient sign method (FGSM) \cite{goodfellow2014explaining} under different perturbation amplitudes $\epsilon \in \{0.00, 0.04, \dots, 0.20\}$ (Figure~\ref{fig:ml-robustness}(c), left). For the ``black-box attack'' case, we train another baseline ANN with $3$ 2D convolutional layers and $1$ fully connected layer on the Fashion-MNIST training set and use FGSM to generate adversarial samples w.r.t. this ANN. Notice that the baseline ANN is shared by all experiment conditions. The adversarial samples are fed to the model of interest for evaluation. Similar to the corruption robustness experiments, we use relative mean adversarial error (rmAE) as the metric
\begin{equation}
\label{eq:methods-rmAE}
\mathrm{rmAE}^f = \frac{\sum_{\epsilon} (\mathrm{acc}^f_{\mathrm{clean}} - \mathrm{acc}^f_\epsilon)}{\sum_{\epsilon} (\mathrm{acc}^{f_b}_{\mathrm{clean}} - \mathrm{acc}^{f_b}_\epsilon)},
\end{equation}
where $f$ is the model of interest, $acc^f_{\mathrm{clean}}$ is the accuracy of $f$ on the clean test set,
$acc^f_\epsilon$ is the accuracy of $f$ on the adversarial test set with perturbation amplitude $\epsilon$,
and $f_b$ refers to the LIF-based SNN.

\subsection*{Details of few-shot learning experiments}
\label{subsec:methods-few-shot-learning}

The miniImageNet benchmark \cite{vinyals2017matching} for few-shot learning consists of $60,000$ images ($84 \times 84$ resolution) of 100 different categories. Among these categories, $64$ classes are used for training, $16$ for validation, and $20$ for testing. We use a SEW ResNet-18 \cite{fang2021sew} as the feature extractor, and construct a prototypical network \cite{snell2017prototypical} on its basis. DendSN replacement is not applied to the first convolution-neuron block, which empirically leads to better accuracies. The classical training pipeline is adopted, where the feature extractor is trained together with a fully connected classification head directly on the entire training set, just like the supervised learning setting. Random resized cropping, color jittering, random horizontal flipping and normalization are applied to augment the training data. Training hyperparameters are listed in Table~\ref{suptab:hyperparameters}.

5-way 1-shot, 5-way 5-shot, 10-way 1-shot, and 10-way 5-shot classification accuracies are adopted as evaluation metrics. Here, an $N$-way $K$-shot classification task refers to a setting where $N$ classes are randomly sampled from the test set, and each class provides $K$ labeled examples as the support set $\mathcal{S}$ for prototype construction. Another set of unlabeled examples from the same $N$ classes forms the query set $\mathcal{Q}$ for evaluation. During the evaluation phase, for each $N$-way $K$-shot task, the trained feature extractor $f$ encodes all samples into an embedding space. For each class $n \in \{1, \dots, N\}$, a prototype vector $\mathbf{p}_n$ is computed as the mean feature of its support samples:
\begin{equation}
\label{eq:methods-fewshot-prototype}
\mathbf{p}_n = \frac{1}{|\mathcal{S}_n|} \sum_{(\mathbf{x}_i, y_i) \in \mathcal{S}_n} f(\mathbf{x}_i),
\end{equation}
where $\mathcal{S}_n$ denotes the subset of $\mathcal{S}$ belonging to class $n$. Each query sample $\mathbf{x}_q \in \mathcal{Q}$ is then classified according to the nearest-neighbor rule, using the Euclidean distance in the embedding space:
\begin{equation}
\label{eq:methods-fewshot-nearest}
\hat{y}_q = \arg\min_{n} \| f(\mathbf{x}_q) - \mathbf{p}_n \|_2^2.
\end{equation}
The classification accuracy is computed by comparing predicted labels $\hat{y}_q$ with the ground-truth labels of all query samples. In our experiment, each class in the query set contains $15$ samples, and the reported accuracy is averaged over $500$ randomly sampled $N$-way $K$-shot tasks.

To gain insights into how DendSNN improves the representational quality underlying few-shot learning, we conduct two visualization analyses based on the features extracted from the trained backbone. For the t-SNE visualization, features are extracted from the query samples of an exemplary 5-way 1-shot task, which contains 256 query instances per class. The features and class prototypes ($D_f=512$ dimensions) are projected onto a 2D manifold using t-SNE \cite{maaten2008tsne} and then plotted. We also visualize the distributions of mean inter-class and intra-class distances across multiple tasks. Specifically, define the center of class $n$ as
\begin{equation}
\label{eq:methods-class-center}
\mathbf{c}_n = \frac{1}{|\mathcal{Q}_n|} \sum_{(\mathbf{x}_i, y_i)\in \mathcal{Q}_n} f(\mathbf{x}_i),
\end{equation}
where $\mathcal{Q}_n$ is the set of query samples belonging to class $n$. $\mathbf{c}_n$ differs from $\mathbf{p}_n$ in that it is computed over all query samples, not support samples. The intra-class distance $d^{\text{intra}}_n$ of class $n \in \{1, \dots, N\}$ is the average Euclidean distance between query features and the class center, while its average over all classes is called the mean intra-class distance $d^{\text{intra}}$.
\begin{align}
d^{\text{intra}}_n &= \frac{1}{|\mathcal{Q}_n|} \sum_{(\mathbf{x}_i, y_i)\in \mathcal{Q}_n} \|f(\mathbf{x}_i) - \mathbf{c}_n\|_2 , \label{eq:methods-intra}
\\
d^{\text{intra}} &= \frac{1}{N} \sum_{n=1}^N d^{\text{intra}}_n . \label{eq:methods-mean-intra}
\end{align}
The mean inter-class distance is defined as the mean pairwise distance between class centers:
\begin{equation}
\label{eq:methods-inter}
d^{\text{inter}} = \frac{2}{N(N-1)} \sum_{1\le i < j \le N} \|\mathbf{c}_i - \mathbf{c}_j\|_2 .
\end{equation}
The inter-to-intra distance ratio $\mu = d^{\text{inter}} / d^{\text{intra}}$ characterizes the relative separability of feature clusters in the embedding space. A larger ratio indicates a better discriminative structure for few-shot classification. Notice that $d^{\text{intra}}$, $d^{\text{inter}}$ and $\mu$ are computed for a given $N$-way $K$-shot task. To enable fair comparison, we randomly sample 500 5-way 1-shot tasks and visualize the distributions of $d^{\text{intra}}$, $d^{\text{inter}}$ and $\mu$ as violin plots in Figure \ref{fig:ml-fewshot}(d).

\bibliography{main}

\section*{Acknowledgements}

The study was funded by the National Natural Science Foundation of China under contracts No. 62425101, No. 62332002, No. 62027804, No.62088102, and the major key project of the Peng Cheng Laboratory (PCL2025A02). We gratefully acknowledge Di Shang and Yuyang Liu for advice on the task-incremental learning experiments, and Kexin Wang, Yuhong Chou and Prof. Yujie Wu for the valuable discussions.

\section*{Author contributions statement}

Y.H. proposed the initial idea, contributed to the design of models and experiments, carried out the experiments, and wrote the paper.
W.F. contributed to the design of models, proposed the acceleration strategies, and revised the paper.
Z.M. contributed to the design of models and experiments, and revised the paper.
G.L. proposed the research direction, contributed to the design of experiments, supervised the progress of the experiments, and revised the paper.
Y.T. contributed to the design of models and experiment, supervised the whole project, and led the writing of this paper.

\section*{Competing interests}

The authors declare no competing interests.

\clearpage
\noindent {\sffamily\bfseries\LARGE Supplementary Materials}
\vspace{1em}
\addcontentsline{toc}{section}{Supplementary Materials}

\renewcommand{\thesection}{S\arabic{section}}
\setcounter{section}{0}
\renewcommand{\thefigure}{S\arabic{figure}}
\setcounter{figure}{0}
\renewcommand{\thetable}{S\arabic{table}}
\setcounter{table}{0}
\renewcommand{\theequation}{S\arabic{equation}}
\setcounter{equation}{0}
\renewcommand{\thealgorithm}{S\arabic{algorithm}} 
\setcounter{algorithm}{0}

\section{Hyperparameter settings}
\label{supsec:hyperparameters}

Table~\ref{suptab:hyperparameters} summarizes the hyperparameter configurations for the experiments. For each task, the learnable model parameters are divided into three groups, which are optimized using distinct configurations as described below.
\begin{itemize}
\item \textbf{Synaptic parameters}: the weights and biases of linear projection layers, including those of batch normalization layers. These parameters are optimized with the initial learning rates (Init. LR) listed in Table~\ref{suptab:hyperparameters} and regularized using an L2 penalty with the corresponding factor (L2 Reg.).
\item \textbf{Branch parameters}: the branch strengths $\mathbf{K}$ of all DendSN layers. Their initial learning rates are obtained by multiplying the base value listed in Table~\ref{suptab:hyperparameters} by their corresponding LR Factor.
\item \textbf{Other neuronal parameters}: all remaining learnable neuronal variables are optimized with the listed initial learning rates and without L2 regularization.
\end{itemize}
All three parameter groups share the same type of optimizer and learning rate scheduler. Here, ``SGD($0.9$)'' denotes the Stochastic Gradient Descent optimizer with a momentum of $0.9$, and ``Cosine'' refers to the cosine annealing learning rate scheduler whose $T_{\text{max}}$ equals the number of training epochs. For the neuronal hyperparameters, $\alpha_{\text{init}}$ is the initial value of the learnable compartmental decay factor $\alpha$, $\beta$ represents the membrane potential decay factor of the LIF soma, and $T_{\text{kernel}}$ indicates the length of the temporal sliding window for in Sliding PSN (used only in L5PC task).

\begin{table}[ht!]
\centering
\caption{
    Hyperparameter settings for the experiments.
}
\label{suptab:hyperparameters}
\begin{tabularx}{\textwidth}{@{}cCCCCCC@{}}
\toprule
 & L5PC & \makecell{(Fashion-)\\MNIST} & \makecell{CIFAR10-\\DVS} & \makecell{Tiny\\ImageNet} & CIFAR-100-C & miniImageNet \\ \midrule
$T$ & 2048 & 4 & 10 & 6 & 6 & 4 \\
Epochs & 1000 & 25 or 100 & 150 & 300 & 300 & 300\\
Batch Size & 16 & 128 & 32 & 128 & 128 & 64 \\
Optimizer & Adam & AdamW & SGD($0.9$) & SGD($0.9$) & SGD($0.9$) & SGD($0.9$) \\
Init. LR & $5\times10^{-4}$ & $1\times10^{-4}$  & 0.1 & 0.1 & 0.1 & 0.1 \\
LR Factor & 1 & 5 & 5 & 1 & 5 & 5 \\
Scheduler & Cosine & None & Cosine & Cosine & Cosine & Cosine \\
L2 Reg. & 0 & 0 & $5\times10^{-4}$ & $5\times10^{-4}$ & $5\times10^{-5}$ & $5\times10^{-5}$ \\
Loss & Equation~\eqref{eq:methods-l5pc-loss} & CE & TET & TET & TET & CE \\
$\alpha_{\text{init}}$ & 0.75 & 0.5 & 0.5 & 0.5 & 0.5 & 0.5\\
$\beta$ & 0.95 & 0.5 & 0.5 & 0.5 & 0.5 & 0.5 \\
$T_{\text{kernel}}$ & 64 & / & / & / & / & /\\
\bottomrule
\end{tabularx}
\end{table}

\section{Dendritic branch activation functions}
\label{supsec:psi}

\begin{figure}[ht!]
\centering
\includegraphics[width=\textwidth]{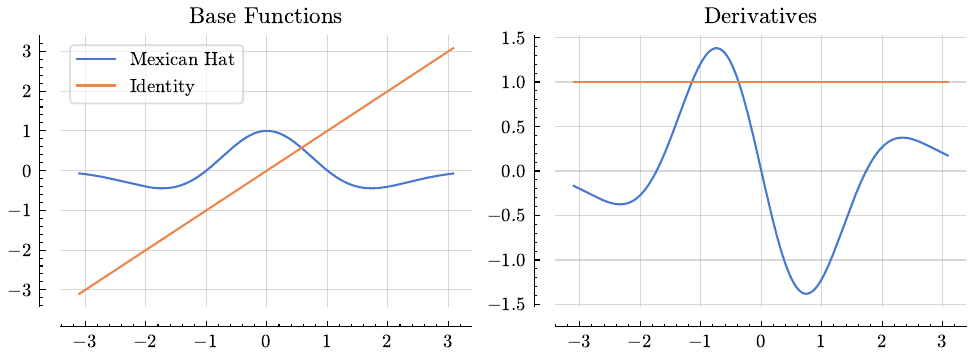}
\caption{Dendritic activation functions $\psi$ explored in this study.}
\label{supfig:psi}
\end{figure}

We explore two types of dendritic nonlinear activation functions in this work, as shown in Figure~\ref{supfig:psi}. Mexican hat (\textbf{MH}) is a bell-shaped function that resembles the receptive field of a simple cell in the visual cortex. It is derived from the second derivative of the Gaussian function. For simplicity, we adopt its standardized form defined in Equation~\eqref{eq:methods-mexican-hat}, where the scale parameter and normalization factor are omitted. We use MH to align our dendritic integration process with the formulation of wavelet neural networks (see Equation~\eqref{eq:results-wnn}). Also, MH can mimic the non-monotonic influence of dendrites on the soma induced by calcium-mediated dendritic activation potentials (dCaAPs) \cite{gidon2020dcaap}. Note that the magnitude of the standard Mexican hat function's derivative is smaller than $1$ for most input values (Figure~\ref{supfig:psi}, right), which may lead to gradient vanishing in deep networks. To alleviate this issue, we propose to simply use identity function (\textbf{Identity}) $\psi(x)=x$ as an alternative. Experiment results show that Identity works better than MH on large networks or complicated tasks (like Tiny ImageNet and miniImageNet), while MH yields better performance on small networks or simple tasks (like Fashion-MNIST).

\section{DendSN's dendrite as a low-pass filter}
\label{supsec:low-pass-filter}

\begin{figure}[ht!]
\centering
\includegraphics[width=0.5\textwidth]{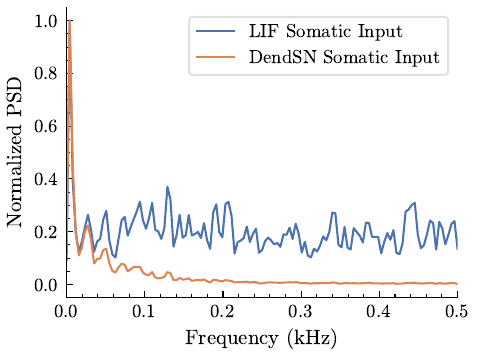}
\caption{
  Frequency domain of LIF and DendSN's somatic input signals. The power spectral density (PSD) is estimated using Welch's method.
}
\label{supfig:low-pass-filter}
\end{figure}

Here we visualize the effect of the proposed dendrite module on the frequency domain for a sample in the L5PC task. We record the somatic input sequence $\{Z[t]\}_{t=1}^{T}$ of a DendSN (Stateful Dendrite, $P=4$, $B=2$, Mexican Hat, $L_2$ aggregation) or a LIF neuron. See Methods for detailed experimental settings. The power spectral density (PSD) of each sequence is estimated using Welch's method and normalized by its maximum value. Figure~\ref{supfig:low-pass-filter} shows that LIF's somatic input contains substantial high-frequency components, whereas the DendSN's somatic input exhibits pronounced attenuation at high frequencies. This demonstrates that the dendrite module in DendSN acts as a low-pass filter, selectively preserving slowly varying temporal features while suppressing high-frequency noise. These visualizations complement Figure~\ref{fig:dendsn} and provide an intuitive view of the feature purification effect induced by the dendrite.

\section{Parameter counts of DendSNNs}
\label{supsec:parameter-counts}

The extra parameter count of a DendSNN compared to its PointSNN counterpart is negligible. To illustrate this, consider the following LIF-based PointSNN subnetwork and its DendSNN counterpart (Stateful Dendrite, LIF soma, $P$ compartments, $B$ branches)
\begin{equation}
\begin{aligned}
\textbf{PointSNN:}\quad  &\text{Conv2d}(\text{in}=C_1, \text{out}=C_2) \to \text{BN}(C_2) \to \text{LIF}(\text{out}=C_2) \\
\to & \text{Conv2d}(\text{in}=C_2, \text{out}=C_3) \to \text{BN}(C_3) \to \text{LIF}(\text{out}=C_3), \\
\textbf{DendSNN:}\quad &\text{Conv2d}(\text{in}=C_1, \text{out}=C_2) \to \text{BN}(C_2) \to \text{DendSN}(\text{out}=C_2/P) \\
\to &\text{Conv2d}(\text{in}=C_2/P, \text{out}=C_3 P) \to \text{BN}(C_3 P) \to \text{DendSN}(\text{out}=C_3).
\label{supeq:subnetwork-parameter}
\end{aligned}
\end{equation}
Also, suppose the 2D convolution kernels have size $k\times k$. Then, the number of learnable parameters for each layer can be summarized as Table~\ref{suptab:parameter-count-calculation}. The total parameter counts are thus
\begin{equation}
\begin{aligned}
\textbf{PointSNN:}\quad & k^2C_1C_2 + k^2C_2C_3 + 3C_2 + 3C_3, \\
\textbf{DendSNN:}\quad & k^2C_1C_2 + k^2C_2C_3 + 4C_2 + 4C_3P + 2C_2B/P + 2C_3B + 2.
\label{supeq:total-parameter-count}
\end{aligned}
\end{equation}
Given that the number of channels $C_1$, $C_2$ and $C_3$ are typically several orders of magnitude larger than $k$, $P$ and $B$, the overall parameter count is dominated by $k^2C_1C_2 + k^2C_2C_3$ (\ie the convolution kernels). Hence, the additional number of parameters introduced by dendrites is negligible. Table \ref{suptab:parameter-count-examples} shows examples of parameter counts for different architectures, vindicating our claim.

\begin{table}[hb!]
\centering
\caption{
    Parameter count for each layer of the two subnetworks in Equation~\eqref{supeq:subnetwork-parameter}.
}
\label{suptab:parameter-count-calculation}
\begin{tabular}{@{}ccccccc@{}}
\toprule
 & Conv2d-1 & BN-1 & Neuron-1 & Conv2d-2 & BN-2 & Neuron-2 \\
\midrule
PointSNN & $k^2 C_1 C_2 + C_2$ & $2C_2$ & $0$ & $k^2 C_2 C_3 + C_3$ & $2C_3$ & $0$ \\
DendSNN & $k^2 C_1 C_2 + C_2$ & $2C_2$ & $1+C_2+2C_2B/P$ & $k^2 C_2 C_3 + C_3P$ & $2C_3P$ & $1+C_3P+2C_3B$  \\
\bottomrule
\end{tabular}
\end{table}

\begin{table}[ht!]
\centering
\caption{
    Parameter count comparison between PointSNNs and DendSNNs of different architectures. PointSNNs adopt LIF neurons. DendSNNs adopt Stateful Dendrite and LIF soma, with $P=4$ and $B=2$. Note that some layers in DendSNNs remain point-neuron based in order to achieve better empirical performance, so the actual total parameter count is even lower than the theoretical estimation made in Equation~\eqref{supeq:total-parameter-count}.
}
\label{suptab:parameter-count-examples}
\begin{tabular}{@{}cccc@{}}
\toprule
\multirow{2}{*}{Architecture} & \multirow{2}{*}{Task} & \multicolumn{2}{c}{Parameter Count $(\times 10^6)$} \\ \cmidrule(l){3-4} 
 &  & PointSNN & DendSNN \\ 
\midrule
FCNet & (Fashion-) MNIST & 5.59 & 5.61 \ExtraPercent{0.36} \\ 
Spiking VGG-11 & CIFAR10-DVS & 9.27 & 9.30 \ExtraPercent{0.32} \\
Spiking VGG-13 & Tiny ImageNet & 9.82 & 9.84 \ExtraPercent{0.20} \\
MS-ResNet18 & CIFAR-100-C & 11.20 & 11.22 \ExtraPercent{0.18} \\
SEW ResNet-18 & miniImageNet & 11.23 & 11.25 \ExtraPercent{0.18} \\
\bottomrule
\end{tabular}
\end{table}

\section{Pseudocode of dendritic branch gating}
\label{supsec:dbg-pseudocode}

Algorithm \ref{supalg:dbg-forward} describes the forward pass of a DendSNN with DBG. Besides the input tensor $\mathbf{X}$, a task index $q$ is also fed into the model as a modulation signal. Algorithm \ref{supalg:dbg-til} show how to use DBG in task incremental learning settings. All we need to do is to replace the forward passes during training and inference with $\mathrm{DBGForward}$. Note that although $\mathrm{DBG}$ randomly samples the masks from a Bernoulli distribution, the masks are fixed once the sampling is done. If $q$, $\rho$, and $\mathbf{K}$ are the same, $\mathrm{DBG}$ will deterministically return the same mask. Therefore, the subnetwork will be correctly activated.

\begin{algorithm}[ht!]
    \caption{$\mathrm{DBGForward}(\mathbf{X}; f, q, \rho)$}
    \label{supalg:dbg-forward}
    \begin{algorithmic}
        \STATE \textbf{Input: } network input $\mathbf{X}$; DendSNN $f$; index of the current task $q$; sparsity factor $\rho$.
        \STATE \textbf{Output: } output of the DendSNN $\widehat{\mathbf{Y}}$.
    \end{algorithmic}
    \begin{algorithmic}[1]
      \STATE \MyComment{modulate $\mathbf{K}$}
      \FOR{each DendSN layer $l$ in $f$} 
        \STATE $\mathbf{K}^{*,l} = \mathbf{K}^l \odot \mathrm{DBG}(q;\mathbf{K}, \rho)$; \quad \MyComment{$\mathbf{K}^l$ is the branch strength matrix of layer $l$}
        \STATE Set $\mathbf{K}^l$ to $\mathbf{K}^{*,l}$;
      \ENDFOR
      \STATE $\widehat{\mathbf{Y}} = f(\mathbf{X})$; \quad \MyComment{forward pass with modulated branch strengths}
      \STATE Return $\widehat{\mathbf{Y}}$.
    \end{algorithmic}
\end{algorithm}

\begin{algorithm}[ht!]
    \caption{Task incremental Learning with dendritic branch gating (DBG)}
    \label{supalg:dbg-til}
    \begin{algorithmic}
        \STATE \textbf{Input: } training and test sets of the $Q$ tasks $\{\mathcal{I}_q^{\text{train}}\}_{q=1}^{Q}, \{\mathcal{I}_q^{\text{test}}\}_{q=1}^{Q}$; loss function $\mathcal{L}$; DendSNN $f$; sparsity factor $\rho$.
        \STATE \textbf{Output: } the trained DendSNN.
    \end{algorithmic}
    \begin{algorithmic}[1]
      \FOR{$q \leftarrow 1$ to $Q$}
        \STATE \MyComment{train $f$ on task $q$}
        \FOR{$(\mathbf{X}, \mathbf{Y}) \in \mathcal{I}_q^{\text{train}}$}
          \STATE $\widehat{\mathbf{Y}} = \mathrm{DBGForward}(\mathbf{X}; f, q, \rho)$
          \STATE Update the parameters of $f$ using BPTT according to the loss $\mathcal{L}(\widehat{\mathbf{Y}}, \mathbf{Y})$
        \STATE Set $\mathbf{K}^l$ to $\mathbf{K}^{*,l}$;
        \ENDFOR
        \STATE \MyComment{test $f$ on task $1$ to $q$}
        \FOR{$q^{*} \leftarrow 1$ to $q$}
          \FOR{$(\mathbf{X}, \mathbf{Y}) \in \mathcal{I}_{q^{*}}^{\text{test}}$}
            \STATE $\widehat{\mathbf{Y}} = \mathrm{DBGForward}(\mathbf{X}; f, q^*, \rho)$
            \STATE Compute the test loss $\mathcal{L}(\widehat{\mathbf{Y}}, \mathbf{Y})$ and other metrics
          \ENDFOR
        \ENDFOR
      \ENDFOR
    \end{algorithmic}
\end{algorithm}

\section{Limitations and future work}
\label{supsec:limitation}

We summarize the limitations of our work from three aspects, and propose future research directions according to them.

\paragraph{Scalability} While we have made significant progress in incorporating dendritic computation into deep SNNs, scaling DendSNNs to larger models and datasets remains a challenge. We have provided early evidence of DendSNNs' scalability by constructing both fully connected and convolutional networks and achieving competitive performance across diverse classification benchmarks. However, there is still a substantial gap between DendSNNs and state-of-the-art deep networks (either SNNs or ANNs) in terms of model size and performance on large-scale tasks. Although DendSNNs can theoretically be extended to other architectures such as Spiking Transformers \cite{zhou2023spikformer,yao2023spikedriventransformer,yao2024meta,zhou2024qkformer} and to models of arbitrary depth, their computational cost and performance on these architectures have not yet been systematically explored. Moreover, while DendSNNs perform well on mid-scale datasets such as TinyImageNet, their generalization ability to high-complexity tasks like ImageNet \cite{deng2009imagenet} remains uncertain. Future work will focus on enhancing the scalability of DendSNNs through improved architectural design and more efficient training methodologies.

\paragraph{Task diversity} Our experiments primarily focus on visual classification tasks involving static images or event-based data. While we have investigated various scenarios including supervised learning, continual learning, robustness against noise and adversarial attacks, as well as few-shot learning, the application scope of DendSNNs can be further broadened. SNNs have already demonstrated promising potential beyond visual classification, particularly in tasks such as object detection and tracking in high-speed scenes. Additionally, dendritic structure endows SNNs with a stronger capacity to process information with rich temporal dynamics \cite{zheng2024temporal,chen2024pmsn,wang2025mmdend}, suggesting their promise for sequence modeling. Moreover, recent studies have successfully employed large-scale SNNs for language modeling and generation. Building upon these advances, DendSNNs may further improve the representational power of SNN-based language models. Our future research will therefore explore extending DendSNNs to a broader spectrum of real-world applications, moving beyond image classification toward domains that demand more intricate temporal and structural processing.

\paragraph{Biological plausibility} While DendSN offers enhanced bio-plausibility compared to point spiking neurons due to its dendritic morphology and nonlinear dynamics, certain limitations remain regarding its biological fidelity. As shown in Figure~\ref{fig:dendsn}(d), although much better than LIF, DendSN still falls short of reproducing the complex activity patterns observed in detailed biophysical models, particularly when the neuron is about to spike. This gap primarily arises from the inevitable simplifications made to ensure low computational cost. Future work may explore alternative formulations of dendritic dynamics that preserve computational efficiency while improving bio-plausibility. Furthermore, the training of DendSNNs currently relies on gradient-based BPTT, which raises concerns about biological feasibility. BPTT requires the transmission of global error signals along a separate backward pathway with weights that are strictly symmetric to the forward ones. However, this assumption is widely regarded as unrealistic in biological neural circuits. To bridge this gap, future research will explore more biologically grounded learning rules for deep DendSNNs. Promising directions include difference target propagation \cite{lee2015difference} that updates parameters using locally generated activity differences, feedback alignment \cite{lillicrap2016random} that relaxes the forward-backward symmetry constraint, and bio-inspired local plasticity rules that leverage dendritic local signals \cite{urbanczik2014learning}.

\end{document}